\documentclass[journal]{IEEEtran} 
\usepackage{amsmath,amsfonts}
\usepackage{algorithmic}
\usepackage{algorithm}
\usepackage{array}
\usepackage[caption=false,font=normalsize,labelfont=sf,textfont=sf]{subfig}
\usepackage{textcomp}
\usepackage{stfloats}
\usepackage{url}
\usepackage{verbatim}
\usepackage{graphicx}
\usepackage{cite}
\usepackage{float}
\usepackage{capt-of}
\usepackage{multirow}
\usepackage{booktabs}
\usepackage{orcidlink}
\usepackage{xcolor}
\usepackage{amssymb}
\usepackage{url} 
\usepackage{longtable}
\usepackage{tabularx} 
\usepackage{booktabs} 
\usepackage{etoolbox}
\usepackage{bbding}
\usepackage{colortbl}
\usepackage{graphicx}
\usepackage{booktabs}
\usepackage{hyperref} 
\usepackage{utfsym}
\usepackage{todonotes} 

\usepackage{diagbox}
\usepackage{makecell}
\usepackage{amsmath}

\definecolor{mytodo}{rgb}{0.8, 0.8, 0.2}

\hypersetup{
    colorlinks=true,
    linkcolor=red,     
    citecolor=blue,    
    urlcolor=purple,     
    linkbordercolor=white 
}

\begin{document}

\title{UniEmoX: Cross-modal Semantic-Guided Large-Scale Pretraining for Universal Scene Emotion Perception}


\author{Chuang~Chen~\orcidlink{0009-0006-0261-4534},
        Xiao~Sun~\orcidlink{0000-0001-9750-7032},~\IEEEmembership{Senior Member,~IEEE,}
        and~Zhi~Liu~\orcidlink{0000-0003-0537-4522},~\IEEEmembership{Senior Member,~IEEE}
       
        \thanks{This work was supported by Special Project of the National Natural Science Foundation of China (62441614), 
        Anhui Province Key R\&D Program (202304a05020068) and General Programmer of the National Natural Science 
        Foundation of China (62376084). (Corresponding authors: Xiao Sun.)}
        \thanks{Chuang Chen is with the School of Artificial Intelligence, Anhui University, Hefei 230601, China, and also with the Institute of Artificial Intelligence, Hefei Comprehensive National Science Center, Hefei 230088, China (e-mail: wa21301027@stu.ahu.edu.cn).}
        \thanks{Xiao Sun is with the Anhui Province Key Laboratory of Affective Computing and Advanced Intelligent Machines, School of Computer Science and Information Engineering, Hefei University of Technology, Hefei 230009, China (e-mail: sunx@hfut.edu.cn).} 
        \thanks{Zhi Liu is with the Department of Computer and Network Engineering, The University of Electro-Communications, Chofu-shi, Tokyo 182-8585, Japan (email: liu@ieee.org).}

}

\markboth{IEEE Transactions on Image Processing}
{Shell \MakeLowercase{\textit{et al.}}: A Sample Article Using IEEEtran.cls for IEEE Journals}

\maketitle

\begin{abstract}
  Visual emotion analysis holds significant research value in both computer vision and psychology. 
  However, existing methods for visual emotion analysis suffer from 
  limited generalizability due to the ambiguity of emotion perception and the diversity of data scenarios. 
  To tackle this issue, we introduce UniEmoX, a cross-modal semantic-guided large-scale pretraining framework. 
  Inspired by psychological research emphasizing the inseparability of the emotional 
  exploration process from the interaction between individuals and their environment, UniEmoX integrates 
  scene-centric and person-centric low-level image spatial structural information, 
  aiming to derive more nuanced and discriminative emotional representations.
  By exploiting the similarity between paired and unpaired image-text samples, 
  UniEmoX distills rich semantic knowledge from the CLIP model to enhance emotional embedding representations 
  more effectively. To the best of our knowledge, this is the first large-scale pretraining framework that integrates psychological theories with contemporary contrastive learning and masked image modeling techniques for emotion analysis across diverse scenarios. 
  Additionally, we develop a visual emotional dataset titled Emo8. 
  Emo8 samples cover a range of domains, including cartoon, natural, realistic, science fiction and advertising cover styles,
  covering nearly all common emotional scenes. 
  Comprehensive experiments conducted on seven benchmark datasets across two downstream tasks 
  validate the effectiveness of UniEmoX. 
  The source code is available at \url{https://github.com/chincharles/u-emo}.
\end{abstract}
\vspace{-5pt} 
\begin{figure}[htbp]
  \begin{center}
  \includegraphics[width=\linewidth]{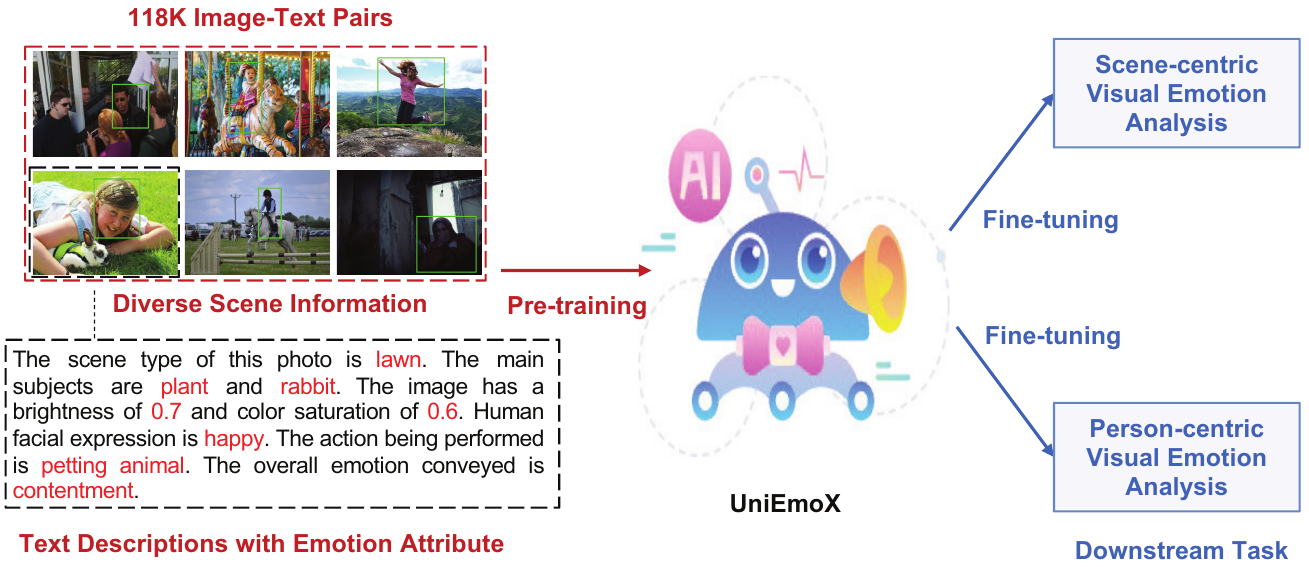}
  \end{center}
  \caption{Overview of UniEmoX: A cross-modal semantic-guided large-scale emotion pretraining framework for downstream visual emotion analysis tasks.}
  \vspace{-15pt} 
  \label{fig:logo}
\end{figure}

\section{Introduction}
\IEEEPARstart{V}{isual} emotion analysis plays a crucial role in various fields such as those of human-computer interaction\cite{4523967}, 
event monitoring\cite{8052551} and recommender system\cite{6362231}.
Currently, most methods\cite{9524517, 9580604, 9846869, kosti2019context, mittal2020emoticon} for visual emotion analysis utilize image classification frameworks and learn 
the mapping from pixel-level data to emotion labels using end-to-end neural networks. However, 
these trained models often demonstrate effectiveness only on specific datasets due to the ambiguity of 
emotional perception~\cite{brainerd2018emotional} and the diversity of data scenarios~\cite{panda2018contemplating}, thus lacking generalizability.

In recent years, large-scale vision models~\cite{he2022masked,radford2021learning} have achieved tremendous success, 
with their performance not yet significantly limited by the capacity of the model. 
Inspired by the paradigm of pre-training large-scale vision models, we can learn rich semantic and visual representations from large-scale datasets. 
Subsequently, fine-tuning on corresponding downstream datasets helps address the issue of 
lack of generalizability in visual emotion analysis methods. We ask: 
How to learn general emotional representations from large-scale complex scenes 
for effective application in various downstream visual emotion analysis tasks?

In the field of computer vision, pre-training paradigms are broadly classified into two primary types: 
supervised pre-training methods~\cite{deng2009imagenet} and unsupervised pre-training methods~\cite{erhan2010does}. 
Supervised pre-training methods involve training on large-scale labeled datasets, 
followed by fine-tuning for specific tasks. 
Self-supervised pretraining methods~\cite{gui2024survey} are a common form of unsupervised learning that does not rely on labeled data. 
They leverage inherent supervisory signals from large-scale unlabeled data by designing auxiliary tasks (i.e., pre-tasks). 
Depending on their design, \IEEEpubidadjcol self-supervised pre-training methods are categorized into several types: 
data augmentation methods (e.g., rotation~\cite{gidaris2018unsupervised}, colorization~\cite{larsson2017colorization}, Jigsaw puzzles~\cite{noroozi2016unsupervised}) 
that generate supervisory signals; 
contrastive learning methods, which maximize similarity among positive samples and minimize it among negative ones; 
and masked image modeling methods, which utilize co-occurrence relationships between image patches as signals. 
Inspired by masked image modeling tasks, 
pixel-level reconstruction of emotionally rich natural images can significantly enhance the understanding 
and modeling of emotions in downstream tasks.

Self-supervised pre-training methods based on contrastive learning, such as CLIP~\cite{radford2021learning}, 
leverage large-scale text-image pairs to learn semantic alignment between text and images. 
CLIP has demonstrated exceptional zero-shot image classification performance. 
Inspired by CLIP's training process, efficiently distilling the semantic knowledge learned by CLIP into the emotional pre-training model 
is crucial for developing a universal emotional pre-training framework.
\begin{figure}[t] 
  \centering
  \begin{minipage}{0.225\textwidth}
      \centering
      \includegraphics[width=\textwidth]{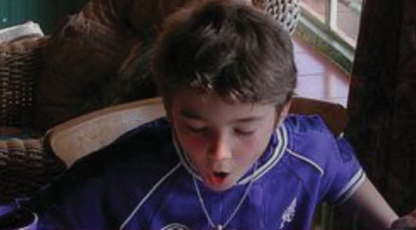}
      \caption{How is this child feeling? Try to recognize his emotional states from the person bounding box, without scene context.}
      \label{fig:subfig1}
  \end{minipage}\hfill
  \begin{minipage}{0.225\textwidth}
      \centering
      \includegraphics[width=\textwidth]{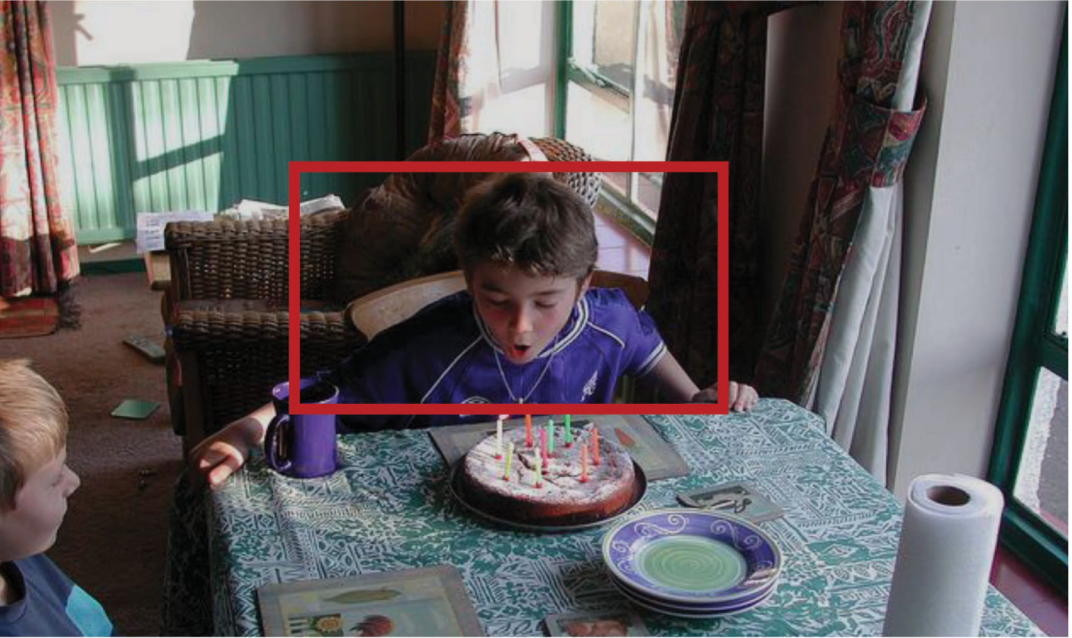}
      \caption{Integrating scene context and individual features for precise emotional content assessment in images.}
      \label{fig:subfig2}
  \end{minipage}
  \vspace{-15pt} 
  \label{fig:overall}
\end{figure}

In psychological research~\cite{barrett2011context, kosti2019context}, interpreting the emotion conveyed by an image involving a person necessitates 
a comprehensive analysis of both scene context and individual features. 
Scene context encompasses elements such as the environment, scene type, 
attributes, surrounding objects, and ongoing events. 
Individual features include facial expressions and body posture. 
For instance, as illustrated in Fig.~\ref{fig:subfig1}, observing only a child makes it challenging 
to accurately determine his emotional state (from his facial expression it seems
he is feeling Surprise). However, with additional scene context (Fig.~\ref{fig:subfig2}), 
such as the child celebrating a birthday and blowing out candles, 
possibly at home with family or friends, we gain a clearer understanding of his expressions and posture. 
This contextual information allows us to infer that the child is likely feeling Happy, Engaged, and Excited. 
Inspired by psychological research and masked image modeling, pixel-level reconstruction of emotionally 
rich natural images, focusing on the interaction between scene and individual features, 
offers valuable insights for how to learn deep emotional representation.

In summary, we propose UniEmoX, a cross-semantic guided multimodal pretraining framework. 
Our aim is to integrate popular pretraining techniques, such as masked image modeling and contrastive learning methods, 
with psychologically validated emotion-related theory prior knowledge to construct a universal emotion representation framework. 
Although our approach may appear as a straightforward integration of existing techniques, the core challenge is ensuring that this integration outperforms 
the use of individual methods in isolation. These difficulties can be summarized in three key aspects. 
\textbf{First}, most existing visual emotion datasets lack a textual modality, 
making it challenging to directly apply CLIP, a multimodal model, in this context. 
Building a new visual emotion dataset with textual descriptions from scratch would require substantial 
time and effort. Efficiently transferring visual-semantic knowledge from CLIP to an emotion pretraining framework remains a challenging problem.
\textbf{Second}, determining how to appropriately incorporate psychological prior knowledge—particularly 
the theory that emotions are inseparable from interactions between humans and their environment—remains a pressing issue. 
\textbf{Finally}, while masked image modeling and contrastive learning techniques are relatively mature individually, 
ensuring stability during training and optimizing performance when combining these two methods 
presents another critical challenge.

\textbf{To address the first difficulty}, we construct 118,000 pairs of emotional images and their corresponding textual descriptions from the EmoSet dataset, 
in which the descriptions are organized using emotion attribute-based natural language logic.
Second, we propose a Contrastive Feature Alignment (CFA) loss and a Semantic-aware Visual Alignment (SVA) loss to effectively distill general visual-semantic knowledge from CLIP.
\textbf{To tackle the second difficulty}, we propose four fusion strategies designed to bridge 
the various semantic gaps between humans and their environment. 
Finally, \textbf{in response to the third difficulty}, 
we introduce a dual-stream multi-loss joint optimization framework grounded in a masked autoencoder, 
incorporating various techniques during the training process, 
including data augmentation and continuous pre-training strategies.

Moreover, unlike the majority of works that focus solely on designing emotion analysis networks 
without addressing the need for high-quality emotion datasets, 
We present Emo8, a high-quality dataset characterized by its diverse sources and extensive variety of emotion labels. 
Although it is not the largest dataset in terms of size, 
Emo8 is distinguished by its comprehensive range of emotional expressions. 
Sourced from the internet, the dataset includes approximately 9 synonyms for each primary emotion label.
We use approximately 80 emotion words as search keywords to obtain 13,200 raw images. 
These images are then rigorously reviewed by multiple individuals to ensure the reliability of the dataset, 
resulting in the high-quality Emo8 dataset, which comprises 8,930 images labeled with 8 distinct emotions.

In summary, our main contributions are as follows:
\begin{itemize}
  \item We develop UniEmoX, a pretraining framework for emotion analysis in universal scenes. 
  UniEmoX effectively addresses the limited generalizability 
  that current visual emotion analysis methods encounter 
  due to the ambiguity of emotion perception and the diversity of data scenarios.
  \item In the development of UniEmoX, we address \textbf{three} primary difficulties. 
  \textbf{First}, to overcome the limitations of textual modalities and the design of contrastive framework, 
  we integrate emotional attributes into texts through natural language logic, 
  constructing 118,000 emotion image-text pairs from the EmoSet dataset.
  We propose a Contrastive Feature Alignment (CFA) loss and a Semantic-aware Visual Alignment (SVA) loss to effectively distill general visual-semantic knowledge from CLIP. 
  \textbf{Second}, To apply psychological theories practically and bridge the semantic gap between humans and their environment, 
  we propose four fusion strategies. 
  \textbf{Lastly}, to improve performance and training stability, 
  we establish a dual-stream, multi-loss joint optimization framework based on masked autoencoding, 
  incorporating various techniques such as data augmentation and 
  continuous pretraining throughout the training process.
  \item We introduce a new dataset titled Emo8, which features diverse sample sources, 
  a wide variety of emotion labels, and delicate emotional expressions. 
  \item We conduct extensive experiments on two downstream tasks: scene-centric visual emotion analysis 
  and person-centric visual emotion analysis, as well as on seven benchmark datasets to validate the effectiveness of our proposed UniEmoX.
\end{itemize}

The rest of this paper is organized as follows.
Section \ref{rw} reviews the related works from scene-centric visual emotion analysis, 
person-centric visual emotion analysis, and large-scale visual model pre-training. 
Section \ref{methodology} details the proposed method. 
Section \ref{dataset} introduces and analyzes the newly constructed Emo8 dataset. 
Section \ref{experiments} presents and discusses the experimental results. 
Conclusions are drawn in Section \ref{conclusion}. 
Finally, Section \ref{limitation} addresses the study's 
limitations and suggests directions for future research.

\section{Related Work}\label{rw} 
In this section, we review existing works related to our UniEmoX from 
scene-centric visual emotion 
analysis, person-centric visual emotion analysis, 
and large-scale visual model pre-training. 
Further details are available in review articles~\cite{9472932, 9591550} on visual emotion analysis.
\subsection{Scene-centric Visual Emotion Analysis}
Scene-centered visual emotion analysis (SVEA) aims to derive a comprehensive emotional summary by 
examining both the background and target information within an image. With advancements in deep learning techniques, 
researchers develope end-to-end classification frameworks to map images to labels. Early works, 
such as the progressive CNN (PCNN)~\cite{you2015robust}
and the multi-level deep representation network (MldrNet)~\cite{rao2020learning}, 
focus on extracting overall image features, 
often neglecting the importance of local regions. 
SOLVER~\cite{9580604}, Stimuli-aware~\cite{9524517}, and MDAN~\cite{9880150} utilize object detection and attention mechanisms 
to extract both global and multi-perspective local features, thereby improving classification performance.
Although current methods achieve strong performance on specific tasks, their complex architectures hinder deployment, 
and ensuring robust generalization across diverse scenarios remains a challenge.

\subsection{Person-centric Visual Emotion Analysis}
Person-centered visual emotion analysis (PVEA) aims to determine the emotional state of specific targets by 
considering both background and target information. Current methods can be broadly classified into two categories. 
The first category primarily focuses on exploring various emotion-related features to 
design sophisticated network structures capable of effectively extracting and integrating these features. 
Early study~\cite{kosti2019context} highlights the significance of contextual information, 
which leads to the development of the EmotiCon~\cite{mittal2020emoticon}. 
EmotiCon introduces types of contexts (gait, facial features, background, and social dynamics)
from a psychological perspective~\cite{chen2023ast, chen2023sta, tawari2013face, kosti2019context, 8576656, WANG2022103679} 
to enrich emotional features. 
Subsequent study~\cite{9373919} developes diverse mechanisms for extracting both foreground and background features, 
integrating multi-dimensional emotional information to enhance the model's expressive capabilities.
The second category focuses on improving classification performance by addressing data background bias and 
developing innovative multi-task learning strategies. 
To mitigate background bias, a post-hoc adjustment method known as the Context Causal Intervention Module (CCIM)~\cite{10636065} 
proposes to counteract the influence of background information on classification outcomes. 
Additionally, the multi-adaptive optimization (MAO) strategy~\cite{hervella2024multi} introduces an adaptive optimization 
technique that dynamically adjusts the contribution of each task to the model parameters, 
thereby enhancing overall classification performance.
However, despite the strong performance of these methods on specific datasets, they, 
similar to SVEA methods, face challenges in achieving widespread deployment in complex scenarios 
due to their intricate structures and limited generalization capabilities.

\subsection{Large-Scale Visual Model Pre-training}
In computer vision, the pre-training paradigm can be divided into two main types: 
supervised pre-training~\cite{deng2009imagenet} and unsupervised pre-training. 
Supervised pre-training involves training on 
large-scale labeled datasets, followed by fine-tuning for specific tasks. 
This study focuses on self-supervised pre-training, a form of unsupervised pre-training that 
does not rely on labeled data. Instead, it extracts supervisory information from large-scale unlabeled data by 
designing auxiliary tasks, known as pretext tasks. Self-supervised pre-training methods can be categorized 
into several types based on different design philosophies: data augmentation-based methods, contrastive learning-based 
methods, and masked image modeling-based methods. Data augmentation-based methods utilize inherent contextual 
relationships between samples, such as spatial structure and the maintenance of local and global consistency, 
to construct supervisory signals. For example, Researchers~\cite{gidaris2018unsupervised} 
train DNNs to learn image representations by recognizing random geometric transformations. 
Additionally, tasks like color prediction~\cite{larsson2017colorization, larsson2017learningrepresentationsautomaticcolorization} 
and jigsaw puzzles~\cite{noroozi2016unsupervised} have been employed as pre-training tasks for self-supervised learning. 
Contrastive learning-based methods aim to maximize the similarity between 
positive samples while minimizing the similarity between negative samples. Starting from the simple instance discrimination 
task, research has evolved to include 
MoCoV3~\cite{xinlei2021empirical}, DINO~\cite{caron2021emerging} and dBOT~\cite{liu2024exploring},
up to the more popular CLIP~\cite{radford2021learning}, showcasing the superiority of contrastive learning. Masked image modeling-based methods leverage the co-occurrence 
relationships between image patches as supervisory signals. For example, BEiT~\cite{bao2021beit} uses the token output of a pre-trained 
tokenizer as its target, while MAE~\cite{he2022masked} directly uses the raw pixels as targets. 
However, the primary aim of these works is to learn low-level visual patterns.
Our goal is to learn general visual emotion representations 
by integrating popular pretraining techniques, such as masked image modeling and contrastive learning, 
with psychology-validated theories of emotion. We address the challenges of combining theoretical methods 
and establish a versatile and effective emotion representation framework, UniEmoX. 
UniEmoX derives rich scene emotion semantics from the large-scale EmoSet dataset 
and improves generalization performance and deployment 
by substituting complex network designs with parameter-detachable ViT structures.

\begin{figure}[htbp]
    \begin{center}
    \includegraphics[width=\linewidth]{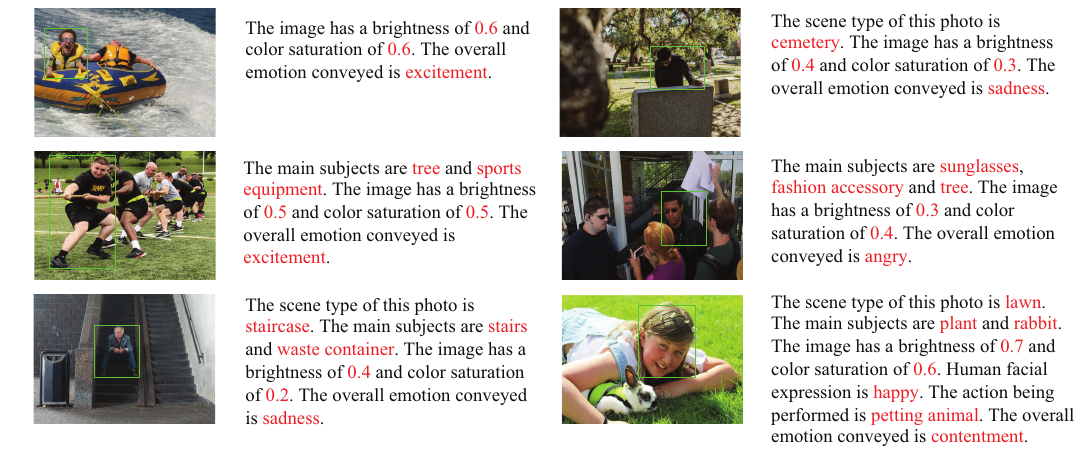}
    \end{center}
    \caption{Partial visual and textual pair examples of constructed training dataset.}
    \vspace{-15pt} 
    \label{fig3_1}
\end{figure}
\begin{figure*}[htbp]
    \centering
    \includegraphics[width=7.0in]{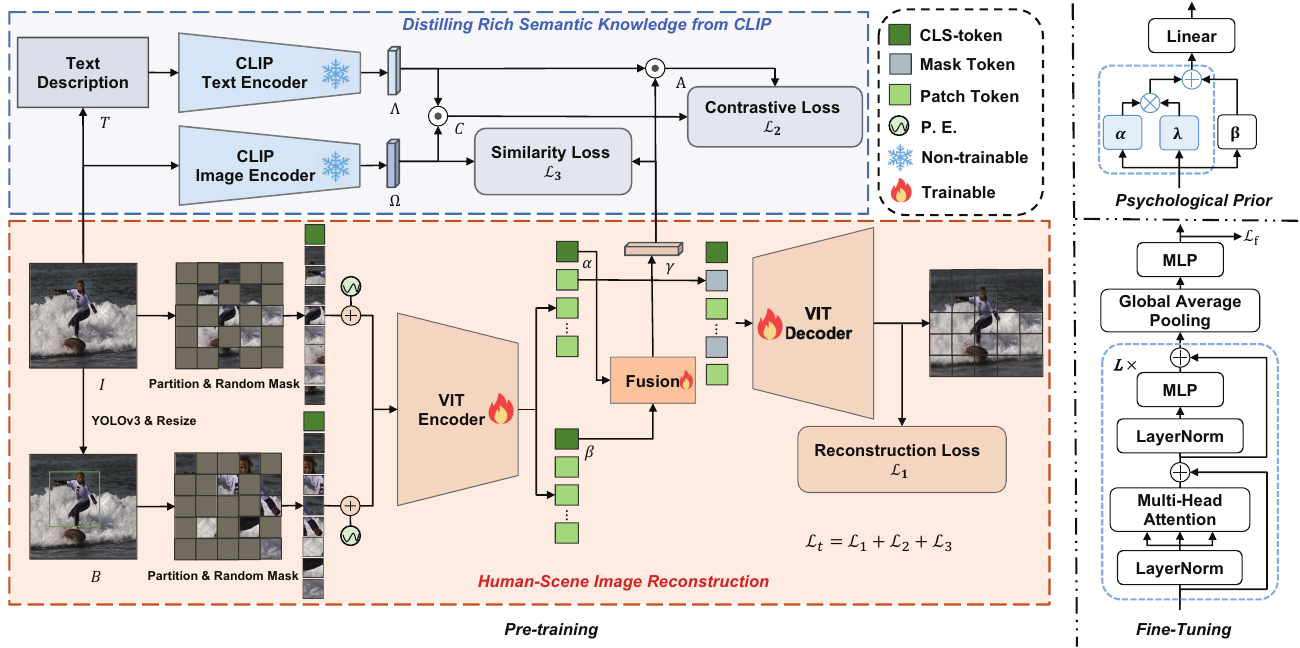}
    \caption{Network architecture of the cross-modal semantic-guided large-scale emotion pretraining framework.}
    \vspace{-15pt} 
    \label{fig:structure}
\end{figure*}

\section{Methodology}\label{methodology} 
We propose UniEmoX, an emotion pretraining framework that synergistically integrates psychological prior knowledge 
with self-supervised learning methods to address the limitations in current visual emotion analysis methods. 
Section~\ref{overview} first introduces the 
foundational architecture and core design principles. 
The architecture comprises three innovative modules: 
(1) Human-Scene Image Reconstruction (Section~\ref{reconstruction}) explicitly models contextual interaction 
features between individuals and scenes; (2) Discriminative Emotional Representation Learning (Section~\ref{discriminative}) bridges 
cross-modal semantic gaps through hierarchical feature fusion; 
and (3) Distilling Rich Semantic Knowledge from CLIP (Section~\ref{distilling}) transfers emotion-relevant knowledge from vision-language models. 
Section~\ref{optimization} optimizes task-specific generalization and cross-domain adaptability. 

\subsection{Overview}\label{overview}
As shown in Fig.~\ref{fig3_1}, the full scene information includes various visual scene maps, 
encompassing the human body and its surrounding environment. 
We define the full scene information map as $I\in \mathbb{R}^{N \times H \times W \times C_{1}} $
and the human position map as $B\in \mathbb{R}^{N \times H \times W \times C_{1}}$. 
UniEmoX is trained on the large-scale emotional dataset EmoSet~\cite{yang2023emoset}, 
which provides multiple label attributes for each sample. 
As illustrated in Fig.~\ref{fig3_1} and Fig.~\ref{fig:structure}, 
we utilize natural language logic to link multiple label attributes, 
thereby constructing text descriptions for the samples, which we refer to as $T$.
As shown in Fig.~\ref{fig:structure}, during the pre-training phase, 
UniEmoX integrates three types of input signals: text descriptions $T$,
full-scene images $I\in \mathbb{R}^{N \times H \times W \times C_{1}} $, and human position maps $B\in \mathbb{R}^{N \times H \times W \times C_{1}}$. 
Motivated by tasks in masked image modeling and the psychological exploration of emotions related to interactions between humans and their environment, 
UniEmoX's backbone processes the full-scene image $I$ and the human position map $B$ as inputs. 
Employing an encoder-decoder architecture for image reconstruction tasks, it integrates learnable fusion features focused on scenes and human bodies. 
This optimization of pixel-level reconstruction enhances the network's ability to effectively represent deep latent layers.
Furthermore, UniEmoX integrates $I$ and corresponding textual descriptions
$T$. Employing knowledge distillation techniques, UniEmoX transfers semantic knowledge from CLIP, 
which enhances its capability to establish a 
universal emotion representation framework with improved generalization.
During the fine-tuning stage for downstream tasks, the parameters from the backbone network are transferred to the VIT architecture, followed by the addition of a linear classification layer tailored to the number of task labels. 
\subsection{Human-Scene Image Reconstruction}\label{reconstruction}
Inspired by Masked Autoencoders (MAE), 
we enhance the model's capability to automatically capture and learn implicit structures 
and contextual information of both individuals and their environments by reconstructing masked regions during pre-training phase. 
This approach improves the model's perception of both the details and global semantics of individuals and their surroundings.

In the reconstruction branch, our UniEmoX follows the processing procedure of MAE, 
performing a series of operations on full-scene images. 
In the reconstruction branch, UniEmoX adopts the MAE processing pipeline, 
applying a series of computational operations to both human and full-scene images. 
For clarity, we outline the computational steps for full-scene images, 
as the same operations are applied to human images. 
These operations include Patch Embedding, Positional Encoding, Random Masking, adding a CLS-token, 
and applying both a Vision Transformer Encoder and Decoder.
Patch Embedding: The full-scene image $I\in \mathbb{R}^{N \times H \times W \times C_{1}}$ is divided into 
$\delta _{1} = \frac{W H}{patch^{2}} $ non-overlapping patches. Each patch is flattened into a one-dimensional vector 
and mapped to the embedding dimension through a linear transformation. 
Consequently, the full-scene image $I$ is transformed into a sequence $S \in \mathbb{R}^{N \times \delta _{1} \times C_{2}}$.
Random Masking: Image patches are masked randomly based on a masking ratio $\mu$, 
leaving $1-\mu$ of the patches unmasked to form sequence $S_{1} \in \mathbb{R}^{N \times \left ( 1 - \mu\right )\delta _{1} \times C_{2}}$.
Positional Encoding: To distinguish between different patches in the sequence, 
each patch is added with a unlearnable positional encoding $P.E. \in \mathbb{R}^{N \times\left ( 1 - \mu\right ) \delta _{1} \times C_{2}}$. 
Consequently, the sequence is represented as $S_{2}=S_{1}+P.E.$.
CLS-token: A learnable CLS-token$\in \mathbb{R}^{N \times 1 \times C_{2}}$ is introduced at the beginning of sequence $S_{2}$. 
This CLS-token's purpose is to integrate global and local image information, 
thereby forming the input sequence $S_{3}\in \mathbb{R}^{N \times\left ( 1 - \mu\right ) \delta _{1}+1 \times C_{2}}$.
Transformer Encoder: The input sequence is processed by a Vision Transformer (ViT) Encoder, 
comprising multiple stacked blocks. Each block includes an attention layer, 
a normalization layer, and a fully connected layer, aiming to extract rich feature representations, 
denoted as $S_{4}\in \mathbb{R}^{N \times\left ( 1 - \mu\right ) \delta _{1}+1 \times C_{2}}$.
Transformer Decoder: After processing $S_{4}$ through a linear layer, 
one part extracts the CLS-token $\alpha \in \mathbb{R}^{N \times 1 \times C_{3}}$, 
obtaining the feature representation of the full-scene image. 
The other part is processed by the decoder to obtain the sequence representation $S'\in \mathbb{R}^{N \times \delta _{1} \times C_{2}}$ of the reconstructed image.

We utilize the design of the loss function from MAE to encourage the model to 
learn meaningful data representation by minimizing the difference 
between the reconstructed image sequence $S'$ of the masked parts and the original image sequence $S$.
The loss function $\mathcal{L}_{1}$ is defined as follows:
\begin{equation}
    \mathcal{L}_{1} = \frac{1}{\left |  M\right | }   {\textstyle \sum_{i\in M}^{}} \left \| S'_{i}-S_{i} \right \|^{2}, 
\end{equation}
where $S_{i}$ represents the original pixel value at position $i$, $S'_{i}$ represents the reconstructed pixel value at position $i$.
$M=\mu \delta _{1}$ is the set of masked positions, $\left | \cdot  \right | $ denotes the size of the set, and $\left \| \cdot  \right \| ^{2}$ indicates the squared error.

As presented in TABLE~\ref{tab:ablation}, incorporating $\mathcal{L}_1$ enhances emotion recognition accuracy by an average of 0.95\% on downstream datasets compared to its absence.
\subsection{Discriminative Emotional Representation Learning}\label{discriminative}
According to psychological prior theories, the cognitive process of emotion is closely tied to the interaction between an 
individual and their environment. However, a natural semantic gap exists between these theories and the design of 
deep learning models. A significant challenge lies in designing effective fusion modules that integrate the feature 
representations of the full-scene image $I$ and the human position map $B$ to derive more nuanced and discriminative emotional 
representations. To address this challenge, we develop four methods, each targeting different degrees of this 
semantic gap.

The full-scene image $I$ is processed by a YOLOv3 detector~\cite{redmon2018yolov3}
and undergoes an image scaling transformation to obtain the human position map 
$B \in \mathbb{R}^{N \times H \times W \times C_{1}}$. The human position map 
$B$ is subjected to similar operations as $I$, passing through the ViT encoder and a linear layer to extract the CLS-token, denoted as $\beta  \in \mathbb{R}^{N \times 1 \times C_{3}}$. 
The feature representations of the body position map $\beta$ and the full-scene image 
$\alpha$ are reduced in dimensionality to  $\mathbb{R}^{N \times C_{3}}$ and $\mathbb{R}^{N \times C_{3}}$, respectively.

To effectively fuse global scene features $\alpha \in \mathbb{R}^{N \times C}$ and human pose features $\beta  \in \mathbb{R}^{N \times C}$, 
 we propose a hierarchical fusion framework with four strategies ($\Gamma_{1}$--$\Gamma_{4}$) tailored to varying degrees 
 of semantic gaps in affective computing.

\subsubsection{Static Channel-weighted Fusion}  
To quantify the necessity of complex fusion strategies, 
 we propose a parameter-free baseline fusion strategy, Static Channel-weighted Fusion, $\Gamma_4$, is defined as follows:
\begin{equation}
\Upsilon = \lambda \odot \alpha + \beta,
\end{equation}
where $\odot$ represents the element-wise (Hadamard) product, and $\lambda \in \mathbb{R}^C$ is a predefined vector of weights. This method serves the following purposes:
\begin{itemize}
    \item Fixed Importance Distribution: Assume that the importance distribution between different semantic features ($\alpha$ and $\beta$) is fixed;  
    \item No Learnable Parameters: Provide a baseline fusion strategy with no learnable parameters, enabling direct evaluation of more complex fusion methods;  
    \item Semantic Gap Evaluation: Quantitatively assess the effectiveness of other strategies in reducing the semantic gap (see TABLE~\Ref{table:fusionResults}).
\end{itemize}

\subsubsection{Dynamic Weighted Fusion}  
To account for the dynamic variation in the importance of scene and human features (e.g., environment-dominant vs. human-dominant emotions), we propose a dynamic weighted fusion method, $\Gamma_{1}$, defined as follows
\begin{equation}
\Upsilon = \operatorname{LayerNorm} \left( \sum_{i=1}^{\kappa} \left( \boldsymbol{\alpha}_i \odot \boldsymbol{\alpha} + \boldsymbol{\beta}_i \odot \boldsymbol{\beta} \right) \right) + \mathbf{b}.
\end{equation}
This method serves the following purposes:  
\begin{itemize}
    \item Simpler Semantic Structures: Designed for datasets where scene clues are sparse, with either scene or human semantics dominating, leading to simpler semantic structures;  
    \item Learnable Weights: Employs multiple sets of learnable weights, $\lambda_i, \mu_i$, to implement $\kappa$ predefined fusion modes, capturing a variety of semantic patterns;  
    \item Adaptive Emotional Mapping: Facilitates the model's learning of adaptive emotional mapping strategies, such as when to prioritize body position or scene features;  
    \item Training Stability: Uses Layer Normalization (LayerNorm) to suppress outliers and improve training stability, particularly when there are significant scale differences between $\alpha$ and $\beta$ via:
    \begin{equation}
    \operatorname{LayerNorm}(\mathbf{x}) = \frac{\mathbf{x} - \mathbb{E}[\mathbf{x}]}{\sqrt{\operatorname{Var}[\mathbf{x}] + \epsilon}} \odot \boldsymbol{\gamma} + \boldsymbol{\beta}.
    \end{equation}
\end{itemize}

\subsubsection{Attention-driven Fusion}  
To handle fine-grained semantic ambiguities, we propose an attention-driven fusion strategy, $\Gamma_{2}$, defined as follows:
\small
\begin{equation}
\Upsilon = \operatorname{Swish} \left( \sum_{i=1}^{\kappa} \left( \operatorname{softmax} \left( \alpha \eta_{i} \right) \odot \alpha + \operatorname{softmax} \left( \beta \xi_{i} \right) \odot \beta \right) \right) + \epsilon.
\end{equation}
\normalsize
This strategy is primarily introduced to:  
\begin{itemize}
    \item Complex Semantic Structures: Address datasets with rich scene clues and complex semantic structures;  
    \item Attention Mechanisms: Utilize multiple attention mechanisms, represented by $\eta_i$ and $\xi_i$, to generate $\kappa$ distinct attention distributions, allowing the model to adaptively select emotional mapping strategies based on sample characteristics.
\end{itemize}

\subsubsection{Gated Attention Fusion}  
To capture additional patterns and handle more complex scenarios, we propose a gated attention fusion method, $\Gamma_{3}$, defined as follows:
\begin{subequations}
    \begin{align}
    \Delta_{i} = \operatorname{softmax}(\alpha \phi_{i}) \odot \alpha + \operatorname{softmax}(\beta \varphi_{i}) \odot \beta,\\
    \Delta_{con} = \operatorname{concat}(\Delta_{1}, \Delta_{2}, \cdots, \Delta_{\kappa}),\\
    \Delta_{m} = \operatorname{ReLU} (\Delta_{con} W_{1} + b_{1}) W_{2} + b_{2},\\
    \Delta_{g} = \Delta_{m} \odot \sigma (\Delta_{m} W_{3} + b_{3}),\\
    \Upsilon = \operatorname{ReLU} (\Delta_{m} \odot \sigma(\Delta_{g})) + \epsilon.\\
\end{align}
\end{subequations}
The introduction of this method serves the following purposes:  
\begin{itemize}
    \item High Noise and Occlusion Handling: Handle complex datasets with high noise and significant occlusion;  
    \item Complementary Alignment: Capture complementary alignment patterns between background and human features through multiple independent parameters, enhancing the model's expressive power;  
    \item Noise Filtering: Employ a gating mechanism ($\sigma$) to dynamically filter noise and retain key cross-modal features.
\end{itemize}

\subsection{Distilling Rich Semantic Knowledge from CLIP}\label{distilling} 
CLIP has demonstrated remarkable potential in cross-modal learning by bridging linguistic and visual modalities. Traditionally, language and vision processing have been treated as separate tasks; 
however, CLIP simultaneously models these modalities through contrastive learning, significantly improving multimodal alignment. 
Building on this success, we propose an emotion-aware pretraining framework that distills knowledge efficiently through two key strategies:
Integration of vision-language contrastive learning paradigms during distillation, implemented via the proposed Contrastive Feature Alignment (CFA) Loss.
Effective transfer of CLIP-acquired visual semantic knowledge to emotion-specific models, achieved through the Semantic-aware Visual Alignment (SVA) Loss.

We construct descriptive texts by naturally linking label attribute from the large-scale emotional dataset EmoSet, 
thereby creating numerous emotion image-text pairs. we input descriptive text $T$ and full-scene image $I$ into CLIP's text encoder and image encoder, 
respectively, obtaining text features $\Lambda \in \mathbb{R}^{N \times C_{3}}$ and visual features $\Omega \in \mathbb{R}^{N \times C_{3}}$.

Given text features $\Lambda \in \mathbb{R}^{N \times C_{3}}$, 
CLIP visual features $\Omega \in \mathbb{R}^{N \times C_{3}}$, 
and visual emotional representations $\Upsilon \in \mathbb{R}^{N \times C_{3}}$, we first perform $\ell_2$-normalization:

\begin{equation}
\Lambda'_i = \frac{\Lambda_i}{\|\Lambda_i\|_2}, \quad
\Omega'_i = \frac{\Omega_i}{\|\Omega_i\|_2}, \quad
\Upsilon'_i = \frac{\Upsilon_i}{\|\Upsilon_i\|_2}, 
\end{equation}
where $\quad \forall i \in \{1,\ldots,N\}.$
This projects features onto a unit hypersphere, eliminating magnitude variations while preserving directional semantics.

Two cross-modal similarity matrices are computed as follows:

\begin{equation}
A = \tau \cdot \Upsilon' \Lambda'^\top,   \\
C = \tau \cdot \Omega' \Lambda'^\top. 
\end{equation}

where $\tau$ is a learnable temperature parameter.

The following content will present the core contributions of this subsection.

\subsubsection{Contrastive Feature Alignment Loss} To achieve efficient knowledge transfer through cross-modal contrastive alignment, we propose a Contrastive Feature Alignment (CFA) Loss $\mathcal{L}_2$, is defined as follows:

\begin{equation}
\mathcal{L}_{2} = 1 - \frac{1}{N} \sum_{i=1}^{N} \left\langle \frac{A_i}{\|A_i\|_2}, \frac{C_i}{\|C_i\|_2} \right\rangle,
\label{eq:loss}
\end{equation}

where $\langle\cdot,\cdot\rangle$ denotes inner product. This design introduces:

\begin{itemize}
    \item Dual Normalization: By applying $\ell_2$-normalization: to each row of $A$ and $C$, 
    the optimization objective shifts from absolute value matching to directional alignment. 
    Compared to traditional knowledge distillation based on KL divergence, 
    this design relaxes constraints to preserve intrinsic geometric properties of heterogeneous features.
    \item Text Bridging: The shared text features $\Lambda'$ establish 
    a unified semantic reference frame, enabling indirect alignment of 
    $\Upsilon'$ and $\Omega'$ while mitigating structural mismatch.
\end{itemize}
As presented in TABLE~\ref{tab:ablation}, incorporating CFA $\mathcal{L}_2$ Loss enhances emotion recognition accuracy by an average of 2.02\% on downstream datasets compared to its absence.

\subsubsection{Semantic-aware Visual Alignment}\label{sec:sva}

To effectively distill the visual semantic knowledge encoded in CLIP's feature representations ($\Omega_n \in \mathbb{R}^{C_{3}}$) 
into our emotion-aware pretraining framework, we propose a Semantic-aware Visual Alignment (SVA) Loss $\mathcal{L}_3$, is defined as follows:

\begin{equation}
\mathcal{L}_3 = 1 - \frac{1}{N} \sum_{n=1}^{N} \frac{\Upsilon_n \cdot \Omega_n}{\|\Upsilon_n\|_2 \|\Omega_n\|_2}.
\label{eq:sva_loss}
\end{equation}

This design facilitates geometric alignment between $\Upsilon_n$ and $\Omega_n$ in the feature space, while intentionally avoiding strict numerical equivalence. Our theoretical rationale stems from two perspectives:

\begin{itemize}
\item Directional Semantic Preservation: 
The cosine similarity metric focuses exclusively on angular alignment rather than vector magnitude matching. This preserves the intrinsic geometry of emotion-specific representations, preventing over-constraint that could force $\Upsilon_n$ to collapse into generic visual semantics. Mathematically, the loss optimizes:

\begin{equation}
\min \theta(\Upsilon_n, \Omega_n) \quad \text{subject to} \quad \|\Upsilon_n\|_2 \neq \|\Omega_n\|_2.
\label{eq:angular_constraint}
\end{equation}

where $\theta$ denotes the inter-feature angle.

\item Hierarchical Knowledge Transfer:
Complete vector alignment would reduce our model to replicating CLIP's visual encoding behavior, thereby limiting its capacity to learn higher-order affective patterns. The magnitude-agnostic design creates a controllable knowledge distillation gap, allowing simultaneous acquisition of 
foundational visual semantics from CLIP and emotion-discriminative features through task-specific learning.

\end{itemize}

As presented in TABLE~\ref{tab:ablation}, incorporating SVA $\mathcal{L}_3$ Loss enhances emotion recognition accuracy by an average of 4.16\% on downstream datasets compared to its absence.

\subsection{The Optimization Objectives in the Fine-Tuning Stage}\label{optimization}
During the fine-tuning phase of downstream tasks,  
we employ soft targets\cite{hinton2015distillingknowledgeneuralnetwork} 
to improve the model's performance. 
Assuming the predicted value of the sample is $x \in \mathbb{R}^{N \times d}$, 
the soft target cross entropy loss $\mathcal{L}_{f}$ can be defined as follows:
\begin{equation}
\mathcal{L}_{\text{soft}} = - \frac{1}{N} \sum_{n=1}^{N} \sum_{i=1}^{d} t_{n,i} \log \left( \sigma(x_{n,i}) \right),
\end{equation}
where \( x_n \in \mathbb{R}^d \) is the output logits (un-normalized prediction) for the \(n\)-th sample, 
\( t_{n,i} \) is the target soft label for the \(n\)-th sample and \(i\)-th class (i.e., the target probability for each class),
\( \sigma(x) \) is the softmax function applied to the logits, producing the predicted class probabilities.

The purpose of this design is to maximize the probability of the correct class while minimizing the probabilities 
of other classes, enabling the model to learn more accurate classification capabilities.

\renewcommand{\footnoterule}{
    \vspace{0.25em} 
    \hrule width \linewidth height 0.4pt 
    \vspace{0.25em} 
}

\makeatletter
\long\def\@makefntext#1{\parindent 1em\noindent
    \hbox to 1.8em{\hss\@makefnmark}%
    {\color{red}\fontsize{6}{10}\selectfont#1}} 
\makeatother

\begin{figure*}[htbp]
    \centering
    \begin{minipage}[b]{0.325\textwidth}
        \centering
        \includegraphics[width=\textwidth]{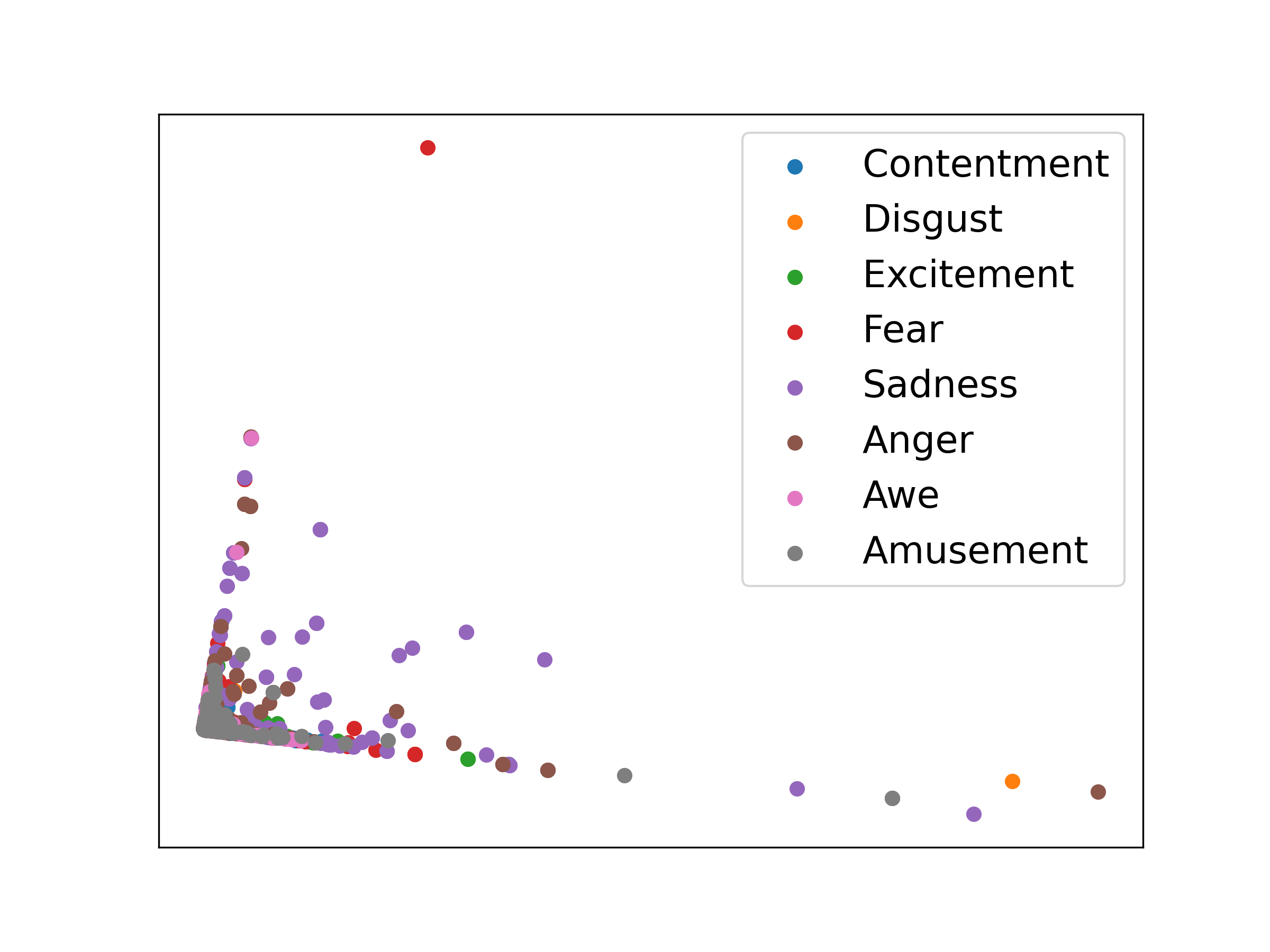}
    \end{minipage}
    \begin{minipage}[b]{0.325\textwidth}
        \centering
        \includegraphics[width=\textwidth]{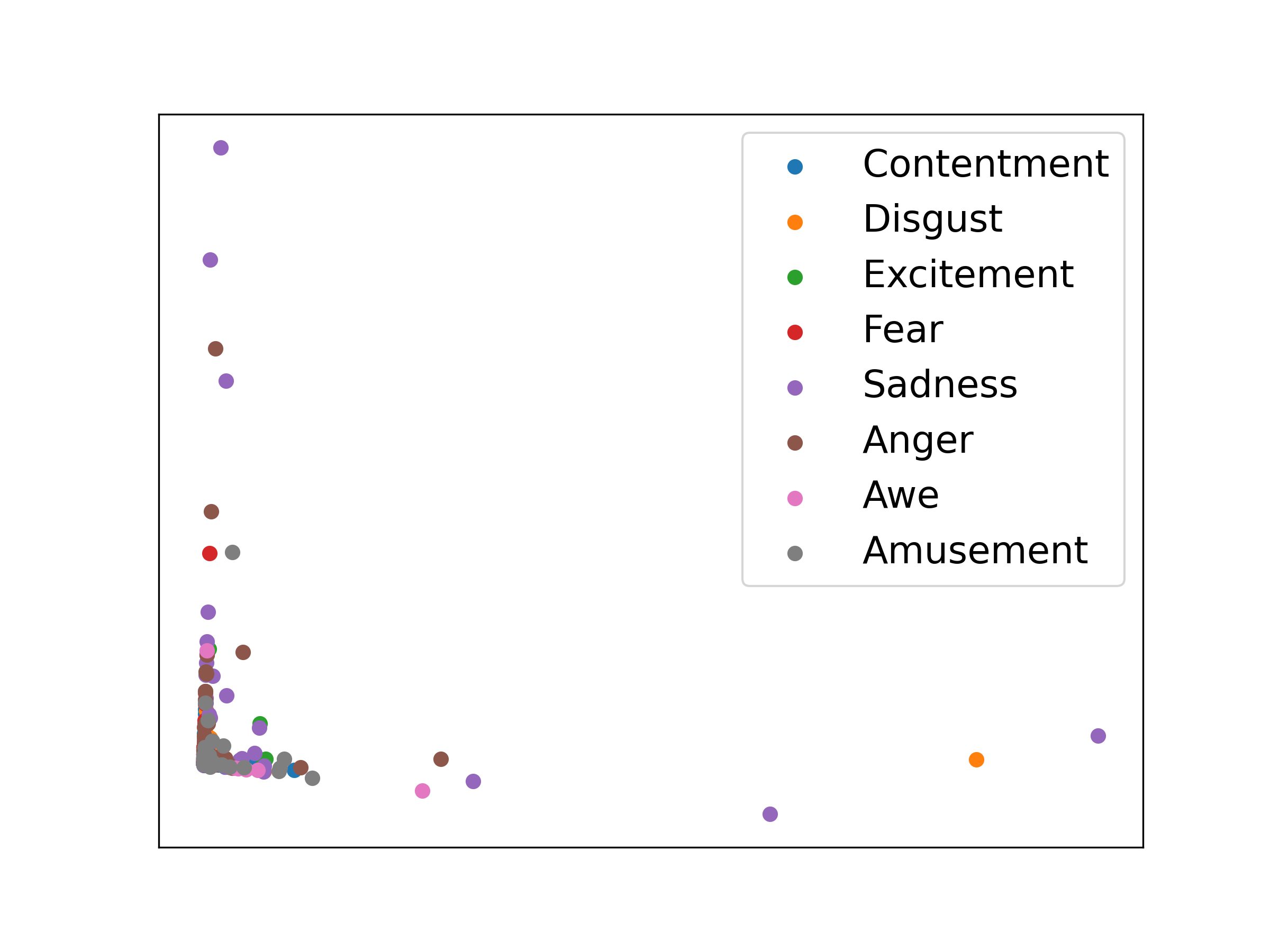}
    \end{minipage}
    \begin{minipage}[b]{0.325\textwidth}
        \centering
        \includegraphics[width=\textwidth]{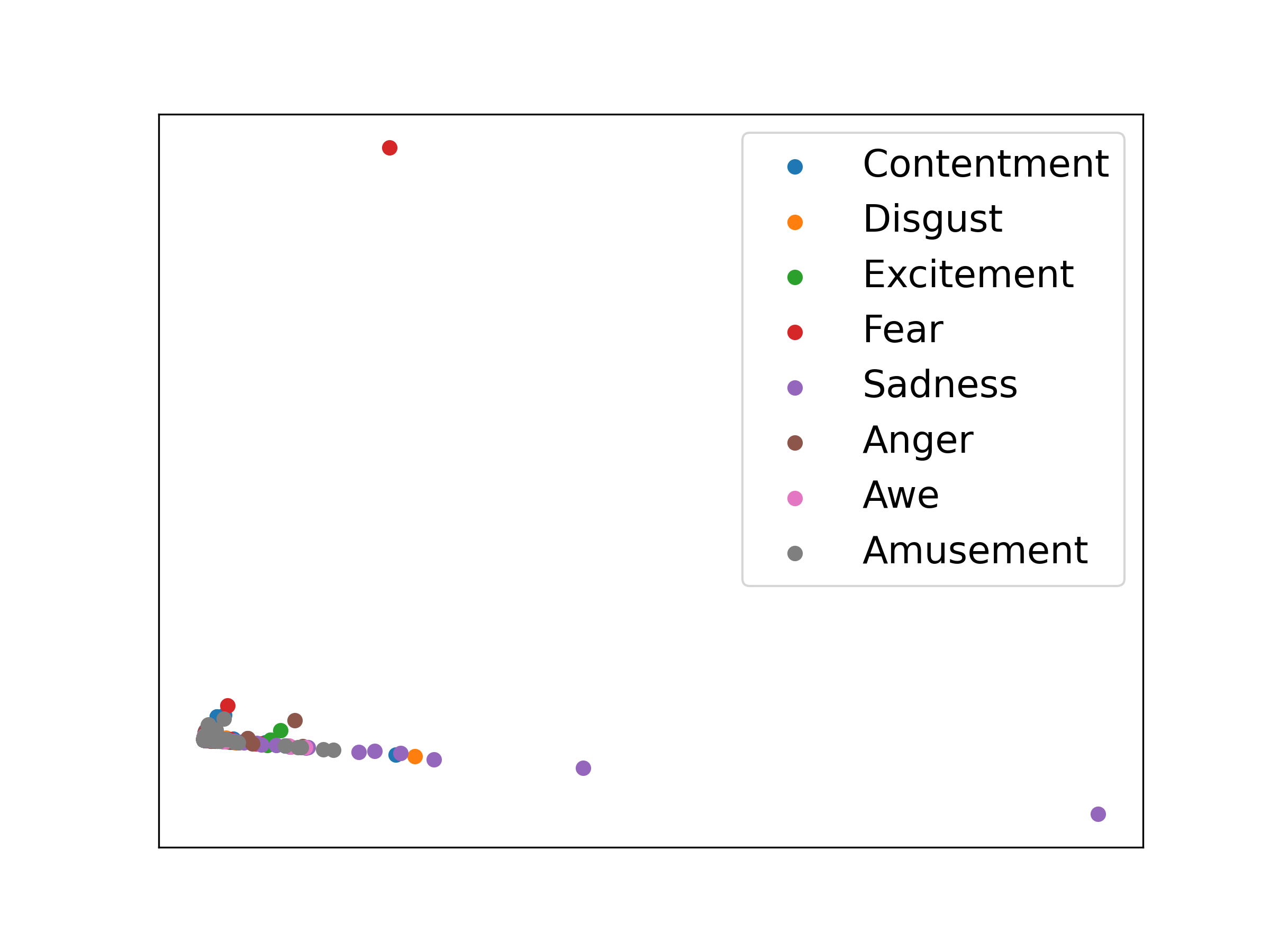}
    \end{minipage}
   \caption{PCA visualization of image sample color features based on color histogram in Emo8, training set, validation set, and test set. 
    Image sample color features from eight emotion labels are sampled and visualized with different colors.}
    \label{fig:pac_visualization}
\end{figure*}
    
\begin{table}
	\centering
	\caption{Search Keywords Corresponding to Each Emotion Category.}\label{tab:keywords}
	\resizebox{1\columnwidth}{!}{
		\begingroup
		\begin{tabular}{m{0.2\columnwidth}p{0.9\columnwidth}} 
			\hline \toprule [0.3 pt]
			
      \multicolumn{1}{l}{\textbf{Emotion}} &   \multicolumn{1}{c}{\textbf{Keyword}}  \\
			\hline 
			Sadness & Sad, Gloomy, Depressed, Unhappy, Melancholy, Mournful, Despondent, Downcast, Sorrowful, Blue \\ 
			Fear & Fear, Dread, Anxiety, Apprehension, Terror, Fright, Panic, Alarm, Unease, Phobia \\ 
			Excitement & Excitement, Thrill, Enthusiasm, Elation, Anticipation, Eagerness, Exhilaration, Fervor, Zeal, Animation \\ 
			Disgust & Disgust, Revulsion, Repulsion, Aversion, Abhorrence, Loathing, Repugnance, Odium, Distaste, Contempt \\ 
			Contentment & Contentment, Satisfaction, Fulfillment, Happiness, Peacefulness, Serenity, Bliss, Ease, Gratification, Tranquility \\ 
			Awe & Awe, Reverence, Admiration, Wonder, Astonishment, Amazement, Respect, Veneration, Dismay, Stupefaction \\ 
			Anger & Anger, Rage, Fury, Wrath, Resentment, Ire, Indignation, Outrage, Annoyance, Hostility \\ 
			Amusement & Amusement, Entertainment, Fun, Enjoyment, Pleasure, Delight, Recreation, Diversion, Pastime, Merriment \\ 
			\hline  \toprule [0.3 pt] 
		\end{tabular}
		\endgroup
	}
	\vspace{-15pt} 
\end{table}

\begin{table}[!htp]
	\centering
	
	\caption{Comparison between visual emotion analysis datasets.}
	\label{tab:datasets}
	\resizebox{1\columnwidth}{!}{
  \begingroup
	\begin{tabular}{lrcc}
		\hline \toprule [0.3 pt]
		\multirow{2}{*}{\bf{Dataset}}&\multirow{2}{*}{\bf{\#Image}} & \bf{Model} & \bf{Image}\\
		&&(\bf{\#category})& \bf{type}\\
		\hline
		IAPSa~\cite{mikels2005emotional}       & 395      & Mikels (8)         & Natural          \\
		Abstract\textsubscript{\textit{MM \textquotesingle 10}}~\cite{machajdik2010affective} & 280      & Mikels (8)         & Abstract         \\
		ArtPhoto\textsubscript{\textit{MM \textquotesingle 10}}~\cite{machajdik2010affective} & 806      & Mikels (8)         & Artistic         \\
		Twitter II\textsubscript{\textit{MM \textquotesingle 13}}~\cite{borth2013large}       & 603      & Sentiment (2)      & Social           \\
		Twitter I\textsubscript{\textit{AAAI \textquotesingle 15}}~\cite{you2015robust}         & 1,269    & Sentiment (2)      & Social           \\
		Emotion6\textsubscript{\textit{CVPR \textquotesingle 15}}~\cite{peng2015mixed}          & 1,980    & Ekman (6)          & Social           \\
		UBE\textsubscript{\textit{ECCV \textquotesingle 18}} ~\cite{panda2018contemplating}     & 3046     & - (6)              & Social           \\
		HECO\textsubscript{\textit{ECCV \textquotesingle 22}}~\cite{yang2022emotion}            & 9,385     & - (8)              & Social           \\
		FI\textsubscript{\textit{AAAI \textquotesingle 16}}~\cite{you2016building}              & 23,308    & Mikels (8)         & Social           \\
		CAER-S\textsubscript{\textit{CVPR \textquotesingle 19}}~\cite{lee2019context}            & 70,000    &  -(7)              & Social        \\
		\hline
		Emo8 (ours)                           & 8,930     &Mikels (8)          & Cartoon, Natural, Realistic, etc.\\
		\hline  \toprule [0.3 pt] 
	\end{tabular}
	\endgroup
}
\vspace{-15pt} 
\end{table}

\section{Emo8 Dataset}\label{dataset} 
\subsection{Dataset Collection}
To construct the Emo8 dataset, we first select the widely-used Mikels model~\cite{mikels2005emotional}, 
which categorizes a broad range of emotions into eight labels: 
amusement, awe, contentment, excitement, anger, disgust, fear, and sadness. 
The first four are positive emotions, while the latter four are negative. As shown in Table~\ref{tab:keywords}, 
we retrieve nine additional synonyms for each of these labels from a dictionary. 
We then input these 80 keywords into the Bing search engine\footnote{\url{https://www.bing.com/}}, 
obtaining approximately 13,200 raw images. 
At this stage, we have 13,200 raw images corresponding to the eight emotion labels. 
Next, we conduct a preliminary screening of the images, mainly removing those that severely mismatch the label meanings. 
Subsequently, we employ two individuals to meticulously validate the emotional content of the filtered images. 
Our criteria are stringent: an image is retained only if both individuals agree that it evoked the corresponding emotion. 
Ultimately, we obtain the Emo8 dataset, comprising 8,930 images across the eight emotion labels. 
Due to space constraints, samples of Emo8 images are provided in Appendix Sections V. 

\subsection{Dataset Statistics}
Compared to other emotion analysis datasets, the Emo8 dataset, though not the largest in terms of image count, 
boasts diverse sample sources, the most comprehensive emotion annotations, 
and delicate emotional expressions. As shown in Table~\ref{tab:datasets}, the Emo8 dataset contains samples in cartoon, 
natural, realistic, science fiction, and advertising cover styles, covering nearly all common emotional scenes.
This ensures the model's generalizability across various contexts during training and validation. 
Additionally, our dataset's adoption of the Mikels model provides a significant advantage in the number of 
annotations compared to datasets using models with fewer annotations, such as Ekman's~\cite{ekman1993facial}. 
Finally, the Emo8 dataset 
offers more delicate emotional expressions due to a three-stage filtering process: image retrieval via the Bing
search engine, initial manual screening, and meticulous manual validation.

Additionally, we conduct an analysis of the segmentation of the Emo8 dataset. 
Ensuring a consistent feature distribution between the validation set and the training set is crucial 
for a good dataset segmentation principle. To validate the quality of our dataset segmentation, 
we perform stratified random splits at a ratio of 8:1:1, generating multiple segmentation schemes. 
We then calculate the color histograms\footnote{\url{https://opencv-tutorial.readthedocs.io/en/latest/histogram/histogram.html}} of the training, validation, and test sets 
for each scheme and perform dimensionality reduction using PCA\footnote{\url{https://scikit-learn.org/stable/modules/generated/sklearn.decomposition.PCA.html}}. 
As shown in Fig.~\ref{fig:pac_visualization}, we select the segmentation scheme with the most consistent distribution between the validation set and the training set 
as the final segmentation scheme.

\definecolor{deemph}{gray}{0.6}
\newcommand{\gc}[1]{\textcolor{deemph}{#1}}
\definecolor{darkgreen}{RGB}{0, 100, 0}
  \begin{table*}[t!]
    
    \tabcolsep=.1em
    \centering
    \caption{
      The performance of twenty methods is compared across two downstream tasks using six datasets. 
      The highest scores in the table are highlighted in \textbf{Bold}, 
      while the second-highest scores are marked in \textcolor{blue}{blue}. 
      Improvement rates are shown in \textcolor{red}{red}, and decline rates in \textcolor{gray}{gray}.
      The symbol ``-'' indicates that the original authors do not provide the relevant data or release the project code.
      The symbol ``\XSolidBrush'' indicates that the model or method is not employed.
      }
    \label{tab:sota}
    \vspace{-.5em}
    \resizebox{1\linewidth}{!}{
    \begin{tabular}{@{}l r r c c c c c c c c c c c @{}}
    \toprule
    \multirow{2}{*}[-0.3em]{Method} 
    & \multicolumn{2}{c}{Metric}
    & \multicolumn{3}{c}{Pre-training epochs (Backbone)}
    & \multirow{2}{*}[-0.3em]{Supervision}
    & \multicolumn{5}{c}{SVEA}
    & \multicolumn{2}{c}{PVEA}  \\ 
    \cmidrule{2-3} \cmidrule{4-6} \cmidrule{8-11} \cmidrule{12-14}
    & Params (M)& FLOPs (G) & ViT-B & ViT-L & CNN &  &\multirow{1}{*}{UBE\cite{panda2018contemplating}} &\multirow{1}{*}{FI\cite{you2016building}} &\multirow{1}{*}{Emotion6\cite{peng2015mixed}} & \multirow{1}{*}{EmoSet\cite{yang2023emoset}} & \multirow{1}{*}{Emo8 (ours)} & \multirow{1}{*}{CAER-S\cite{lee2019context}}& \multirow{1}{*}{HECO\cite{yang2022emotion}}\\
    \midrule
    \multicolumn{2}{l}{\textit{\gc{Scratch}}}\\
    Scratch, ViT\textsubscript{\textit{ICLR \textquotesingle 21}}~\cite{dosovitskiy2020image} 
    &303.31 & 61.60 &  \usym{2717}   &   \usym{2717}    &   \usym{2717}      & \usym{2717}
    &31.89  & 37.44 &32.09  & 35.32 & 31.89 &22.01  & 31.57\\
    \midrule
    
    \multicolumn{2}{l}{\textit{\gc{Supervised models}}}\\
    \href{https://pytorch.org/vision/main/models/generated/torchvision.models.vgg16.html}{VGG-16}\textsubscript{\textit{ICLR \textquotesingle 15}}~\cite{simonyan2014very}            
    &134.29 & 15.47 & \usym{2717}  &  \usym{2717}     & 100+    & Label
    &59.58  & 59.84 &57.46 & 64.97 & 58.67 & 14.29 & 38.09 \\
      
    \href{https://pytorch.org/vision/main/models/generated/torchvision.models.resnet50.html}{ResNet-50}\textsubscript{\textit{CVPR \textquotesingle 16}}~\cite{he2016deep}  
    &23.52  &4.14   &  \usym{2717} &  \usym{2717}     &  100+    & Label      
    &60.78  &57.00  &56.97 & 65.52 & 57.00 & 76.05 & 39.21 \\
      
    \href{https://pytorch.org/vision/main/models/generated/torchvision.models.densenet121.html}{DenseNet-121}\textsubscript{\textit{CVPR \textquotesingle 17}}~\cite{huang2017densely} 
    &6.96   & 2.91  & \usym{2717}  & \usym{2717}      &  100+     & Label
    &60.33  & 60.24 & 58.21 & 64.33 & 55.44 & 77.49 &40.02 \\
  
    \href{https://pypi.org/project/timm/}{ViT}\textsubscript{\textit{ICLR \textquotesingle 21}}~\cite{dosovitskiy2020image}     
    &303.31 & 61.60 &  \usym{2717} &    \usym{2717}   &  100+   & Label
    &38.47  & 53.82 & 63.68 & 75.74 & 54.89 &77.17  &34.62 \\
        
    \midrule
    \multicolumn{2}{l}{\textit{\gc{Self-supervised models}}}\\
    
    \href{https://github.com/facebookresearch/moco-v3/tree/main}{MoCov3}\textsubscript{\textit{ICCV \textquotesingle 21}}~\cite{xinlei2021empirical} 
    &215.68 & 141  & 300 &   \usym{2717}     & \usym{2717}     &Pixel   
    &61.08  & 39.19&61.44 & 55.03    & 57.56  &39.70 &44.71 \\
  
    \href{https://github.com/facebookresearch/dino/tree/main}{DINO}\textsubscript{\textit{ICCV \textquotesingle 21}}~\cite{caron2021emerging}     
    &108.94 & 60.94 & 1600 &  \usym{2717}    &  \usym{2717}    &Feature
    &67.67  &65.89  & 59.20& 70.67  &64.56 &54.25 &37.98 \\
  
    \href{https://github.com/microsoft/unilm/tree/master/beit}{BEIT}\textsubscript{\textit{ICLR \textquotesingle 22}}~\cite{bao2021beit}      
    &91.97  & 18.054 & 800 &   \usym{2717}   &   \usym{2717}  & DALL-E
    &35.30  & 28.10  &31.80&  28.30 &28.90 &15.90 & 29.10 \\
  
    \href{https://github.com/facebookresearch/mae/tree/main}{MAE}\textsubscript{\textit{CVPR \textquotesingle 22}}~\cite{he2022masked}      
    &329.54 & 20.78 &  \usym{2717}   & 1600  &  \usym{2717}    &Pixel
    &35.92  & 59.07 & 41.29 & 74.96  & 77.89 &21.56 &33.60 \\
      
    \href{https://github.com/microsoft/SimMIM?tab=readme-ov-file}{SimMIM}\textsubscript{\textit{CVPR \textquotesingle 22}}~\cite{9880205} 
    &86.26  & 17.70 & 800 &  \usym{2717}    &  \usym{2717}    &Pixel  
    &82.78  & 67.18 & 58.71& 77.84  &73.33 &84.09 &45.83\\
  
    \href{https://github.com/OliverRensu/TinyMIM?tab=readme-ov-file}{TinyMIM}\textsubscript{\textit{CVPR \textquotesingle 23}}~\cite{10204508}
    &85.80 & 16.53 & 300 &    \usym{2717}   &   \usym{2717}   &Feature
    &83.38 & 71.49 & 60.95& 77.98  &77.67  &86.75 &47.15\\
    
    \href{https://github.com/OpenGVLab/Siamese-Image-Modeling}{SiameseIM}\textsubscript{\textit{CVPR \textquotesingle 23}}~\cite{10205066} 
    &231.25& 37.65 &  1600&  \usym{2717}     &   \usym{2717}   &Feature
    &83.83 & 70.63 &  65.67& 76.39   & 80.00 &88.20 &46.74\\

    \href{https://github.com/Sense-X/MixMIM}{MixMAE}\textsubscript{\textit{CVPR \textquotesingle 23}}~\cite{10204021} 
    &223.47 & 38.56 &  \usym{2717}   & 600* &   \usym{2717}   &Pixel  
    &\textcolor{blue}{83.83}  & 69.99 & 62.69& \textcolor{blue}{80.19} & 73.67 &83.12 &\textbf{48.37}\\

    \href{https://github.com/liuxingbin/dbot/tree/main}{dBOT}\textsubscript{\textit{ICLR \textquotesingle 24}}~\cite{liu2024exploring} 
    &329.67 & 82.40 &  \usym{2717}   &1600  &   \usym{2717}     &Feature
    &84.43  & 71.25 & 60.70&  79.75  &79.22 &  88.55 & \textcolor{blue}{47.86}\\

    \href{https://github.com/Atten4Vis/CAE/tree/master}{CAE}\textsubscript{\textit{IJCV \textquotesingle 24}}~\cite{chen2024context} 
    &234.79 & 24.246 &  \usym{2717}  & 1600  &  \usym{2717}    &Pixel + Feat
    &82.93  & 68.19  & \textcolor{blue}{64.18} &\textbf{80.19}  &\textcolor{blue}{80.78}  &90.78 &45.32\\

    \midrule
    \multicolumn{2}{l}{\textit{\gc{Previous SOTA in VEA}}}\\
    SOLVER\textsubscript{\textit{TIP '21}}~\cite{9580604} 
    &-   &-     &\multicolumn{3}{c}{ResNet-18, ResNet-50}  &Label
    &-    &\textcolor{blue}{72.33} &62.12 &   &-&-&-\\

    Stimuli-aware\textsubscript{\textit{TIP '21}}~\cite{9524517}
    &-   &-     &\multicolumn{3}{c}{ResNet-50, ResNet-101} &Label
    &-    &\textbf{72.42} &61.62 &78.40   &-&-&-\\
    
    MDAN\textsubscript{\textit{CVPR '22}}~\cite{9880150}
    &48.79 &101.4 &\multicolumn{3}{c}{ResNet-101}   &Label
    &-    &-     &61.66 &75.75    &-&-&-\\

    EAP\textsubscript{\textit{IJCAI '22}}~\cite{zhang2022visual} 
    &-&-&\multicolumn{3}{c}{BERT} &Label
    &81.77&-    &-  & &-&-&-\\

    RRLA\textsubscript{\textit{TAFFC '23}}~\cite{9373919}
    &-&-&\multicolumn{3}{c}{ResNet-50}&Label
    &-   &-  &-   &   &-&84.82&-\\

    EmotiCon + CCIM\textsubscript{\textit{TPAMI '24}}~\cite{10636065}
    &-&-&\multicolumn{3}{c}{ResNet-50}&Label
    &-&-&-   &  &-&\textcolor{blue}{91.17}&-\\

    \midrule

    \href{https://github.com/chincharles/u-emo}{UniEmoX} 
    &329.54 & 20.78& & 100 or 200& &CLIP B/16 + Pixel
    &\textbf{84.43} \textcolor{red}{$\uparrow$ 0.6} 
    &  69.33\textcolor{gray}{$\downarrow $ 3.09} 
    & \textbf{65.67} \textcolor{red}{$\uparrow$ 1.49} 
    & 78.71 \textcolor{gray}{$\downarrow $ 1.48}
    &\textbf{81.33} \textcolor{red}{$\uparrow$ 0.55} 
    &\textbf{92.09} \textcolor{red}{$\uparrow$ 0.92} 
    & 47.25\textcolor{gray}{$\downarrow $ 1.12}\\

    \bottomrule
    \end{tabular}
    }
    \vspace{-15pt} 
    \end{table*}

\definecolor{deemph}{gray}{0.6}
\definecolor{darkgreen}{RGB}{0, 100, 0}
\begin{table*}[t!]
	\centering
	\tabcolsep=1.2em
	\centering
	\caption{Ablation studies on three components: human-environment image reconstruction, 
    semantic knowledge distillation from CLIP, and incorporation of psychological prior in human-environment feature fusion.}
	\label{tab:ablation}  
    \vspace{-.5em}
	\resizebox{1\linewidth}{!}{

	\begin{tabular}{@{}l c c c c c c c c c c c c @{}}
		\toprule
		\multirow{2}{*}[-0.3em]{Component} 
        & \multicolumn{3}{c}{Image Reconstruction} 
        & \multicolumn{6}{c}{Semantic Knowledge} 
        & \multicolumn{3}{c}{Psychological Prior} \\
		\cmidrule{2-4} \cmidrule{5-10}\cmidrule{11-13}
		
        & w/o $\mathcal{L}_{1}$ & w/ $\mathcal{L}_{1}$ & $\Delta$ & w/o $\mathcal{L}_{2}$ & w/ $\mathcal{L}_{2}$ & $\Delta$  & w/o $\mathcal{L}_{3}$ & w/ $\mathcal{L}_{3}$ & $\Delta$ & w/o fusion & w/ fusion & $\Delta$    \\
		\midrule
		UBE      & 83.38  & \textbf{84.43} & \textcolor{red}{$\uparrow$ 1.05}  & \textbf{84.58}  & 84.43  & \textcolor{gray}{$\downarrow$ 0.15}  & 82.64 & \textbf{84.43} & \textcolor{red}{$\uparrow$ 1.79}  & 82.19  & \textbf{84.43} & \textcolor{red}{$\uparrow$ 2.24} \\
		FI       & 67.99  & \textbf{69.33} & \textcolor{red}{$\uparrow$ 1.34}  & 68.82  & \textbf{69.33}  & \textcolor{red}{$\uparrow$ 0.51}  & 66.81 & \textbf{69.33} & \textcolor{red}{$\uparrow$ 2.52}  & \textbf{69.56}  & 69.33 & \textcolor{gray}{$\downarrow$ 0.23} \\
		Emotion6 & 64.68  & \textbf{65.67} & \textcolor{red}{$\uparrow$ 0.99}  & 63.18  & \textbf{65.67}  & \textcolor{red}{$\uparrow$ 2.49}  & 60.44 & \textbf{65.67} & \textcolor{red}{$\uparrow$ 5.23}  & 59.45  & \textbf{65.67} & \textcolor{red}{$\uparrow$ 6.22} \\
		EmoSet   & 77.75  &\textbf{78.71}    & \textcolor{red}{$\uparrow$ 0.96}  & 77.89  & \textbf{78.71}    & \textcolor{red}{$\uparrow$ 0.82}  & 75.77 &\textbf{78.71}    & \textcolor{red}{$\uparrow$ 2.94}  & 78.63  & \textbf{78.71}   & \textcolor{red}{$\uparrow$ 0.08} \\
		Emo8     & 80.44  & \textbf{81.33} & \textcolor{red}{$\uparrow$ 0.89}  & 80.44  & \textbf{81.33}  & \textcolor{red}{$\uparrow$ 0.89}  & 78.11 & \textbf{81.33} & \textcolor{red}{$\uparrow$ 3.22}  & 80.00  & \textbf{81.33} & \textcolor{red}{$\uparrow$ 1.33} \\
		CAER-S   & \textbf{92.68}  & 92.09 & \textcolor{gray}{$\downarrow$ 0.59}  & \textbf{92.69}  & 92.09  & \textcolor{gray}{$\downarrow$ 0.60}  & 90.39 & \textbf{92.09} & \textcolor{red}{$\uparrow$ 1.70}  & \textbf{92.45}  & 92.09 & \textcolor{gray}{$\downarrow$ 0.36} \\
		HECO     & 46.23  & \textbf{47.25} & \textcolor{red}{$\uparrow$ 1.02}  & 37.07  & \textbf{47.25}  & \textcolor{red}{$\uparrow$ 10.18} & 35.54 & \textbf{47.25} & \textcolor{red}{$\uparrow$ 11.71} & 38.70  & \textbf{47.25} & \textcolor{red}{$\uparrow$ 8.55} \\
        \midrule
		Average  & --  & --  & \textcolor{red}{$\uparrow$ 0.95}  & --  & --  & \textcolor{red}{$\uparrow$ 2.02}  & --  & --  & \textcolor{red}{$\uparrow$ 4.16}  & --  & --  & \textcolor{red}{$\uparrow$ 2.69} \\
		\bottomrule
	\end{tabular}
    }
\end{table*}

\begin{table}
    \centering 
    \tabcolsep=.3em
    \caption{Impact of varied proportions of masked tokens on accuracy across diverse datasets.} 
    \label{table:ratioMaskResults} 
   
    \resizebox{1\linewidth}{!}{
        \begingroup
        \begin{tabular}{c|ccccc|cc} 
        \hline  \toprule [0.3 pt] 
        \multirow{2}{*}{\raggedright \textbf{Ratio of Masked Token}} &
        \multicolumn{5}{c|}{\textbf{SVEA}} &
        \multicolumn{2}{c}{\textbf{PVEA}} \\
        ~ &UBE & FI & Emotion6 & EmoSet & Emo8 (ours) & CAER-S& HECO  \\ 
    
        \hline 
        0.25 &75.45          &61.27          &  56.22           &  73.58        &      77.11          &92.67          &33.71 \\
        0.50 &\textbf{75.75} &\textbf{61.30} &  \textbf{59.45}  & \textbf{73.85}&      \textbf{78.00} &92.60          &36.76 \\
        0.75 &71.41          &60.30          &  53.98           &  72.43        &      75.33          &\textbf{92.84} &\textbf{40.63}\\
        0.85 &71.41          &60.44          &  48.01           &  70.61        &      75.78          &92.47          &37.88 \\
        \hline  \toprule [0.3 pt] 
        \end{tabular}
    \endgroup
    }

    \vspace{-10pt} 
\end{table} 
\begin{table}
    \centering 
    \scriptsize	
    \caption{
        Impact of different fusion strategies on accuracy across diverse datasets.} 
    \label{table:fusionResults} 
   
    \resizebox{1\columnwidth}{!}{
        \begingroup
        \scriptsize	
        \begin{tabular}{c|ccccc|cc} 
        \hline  \toprule [0.3 pt] 
        \multirow{2}{*}{\raggedright \textbf{Component}} &
        \multicolumn{5}{c|}{\textbf{SVEA}} &
        \multicolumn{2}{c}{\textbf{PVEA}} \\
        ~ &UBE & FI & Emotion6 & EmoSet & Emo8 (ours) & CAER-S & HECO  \\ 
        \hline 
        $\Gamma_{1}$ &\textbf{76.80} &61.45          & 54.98         &   72.14   &\textbf{77.22} &90.88          &35.74 \\
        $\Gamma_{2}$ &76.35          &\textbf{62.57} & \textbf{54.98}         &   72.73   &76.22          &90.15          &\textbf{42.16} \\
        $\Gamma_{3}$ &76.80          &62.36          & 54.23         &   \textbf{72.78}   &76.56          &89.51          &37.37 \\
        $\Gamma_{4}$ &71.41          &60.30          & 53.98         &   72.43   &75.33          &\textbf{92.84} &40.63\\
    
        \hline  \toprule [0.3 pt] 
        \end{tabular}
    \endgroup
    }
    \vspace{-15pt} 
\end{table}

\begin{table}
    \centering 
    \scriptsize	
    \caption{``w/ labels'' indicates that the text descriptions include manually annotated emotional 
    information during pretraining, whereas ``w/o labels'' indicates the absence of such annotations.} 
    \label{table:emotion_label} 
   
    \resizebox{1\columnwidth}{!}{
        \begingroup
        \scriptsize	
        \begin{tabular}{c|ccccc|cc} 
        \hline  \toprule [0.3 pt] 
        \multirow{2}{*}{\raggedright ~} &
        \multicolumn{5}{c|}{\textbf{SVEA}} &
        \multicolumn{2}{c}{\textbf{PVEA}} \\
        ~ &UBE & FI & Emotion6 & EmoSet & Emo8 (ours) & CAER-S& HECO  \\ 
    
        \hline 
       
        w/ labels 
        & 84.43 & 69.33 & 65.67 & 78.71  & 81.33 & 92.09 & 47.25\\

        w/o labels
        & 85.03 & 67.67 & 65.67 & 77.98  & 80.78 & 92.34 & 45.62 \\
     
        Difference 
        & \textcolor{red}{$\uparrow$ 0.6} 
        & \textcolor{gray}{$\downarrow$ 1.66} 
        &  -
        & \textcolor{gray}{$\downarrow$ 0.73} 
        & \textcolor{gray}{$\downarrow$ 0.55} 
        & \textcolor{red}{$\uparrow$ 0.25} 
        & \textcolor{gray}{$\downarrow$ 1.63} \\

        \hline  \toprule [0.3 pt] 
        \end{tabular}
    \endgroup
    }

    \vspace{-15pt} 
\end{table} 
\section{Experiments}\label{experiments} 
\subsection{Datasets}
By linking the emotional label attributes of the EmoSet\cite{yang2023emoset} dataset through natural language logic, 
we construct 118,000 emotion text-image pairs for pretraining UniEmoX. 
To evaluate the effectiveness and generalization ability of UniEmoX, 
we conduct comparison experiments, ablation studies, and visual analyses on five public benchmark datasets: CAER-S\cite{lee2019context}, HECO\cite{yang2022emotion}, UBE\cite{panda2018contemplating}, Emotion6~\cite{peng2015mixed}, EmoSet and FI\cite{you2016building} as well as our Emo8 dataset.
Due to space constraints, descriptions of downstream tasks and their corresponding datasets are provided in Appendix Section I.

\subsection{Evaluation Metrics and Implementation Details}
In this work, we adopt $Accuracy$ as the evaluation metric. 
$Accuracy$ denotes the ratio of correctly classified samples by the model, calculated as follows
\begin{equation}
    Accuracy = \frac{TP + TN}{TD}, 
\end{equation}
where $TP$, $TN$, $TD$ are the number of true positives, true negatives, and total data, respectively.

Due to space constraints, detailed information on experimental parameter configurations and model efficiency analyses is provided in Appendix Sections II and III.
\begin{figure*}[htbp]
  \centering
  \begin{minipage}[b]{0.329\textwidth}
      \centering
      \includegraphics[width=\linewidth]{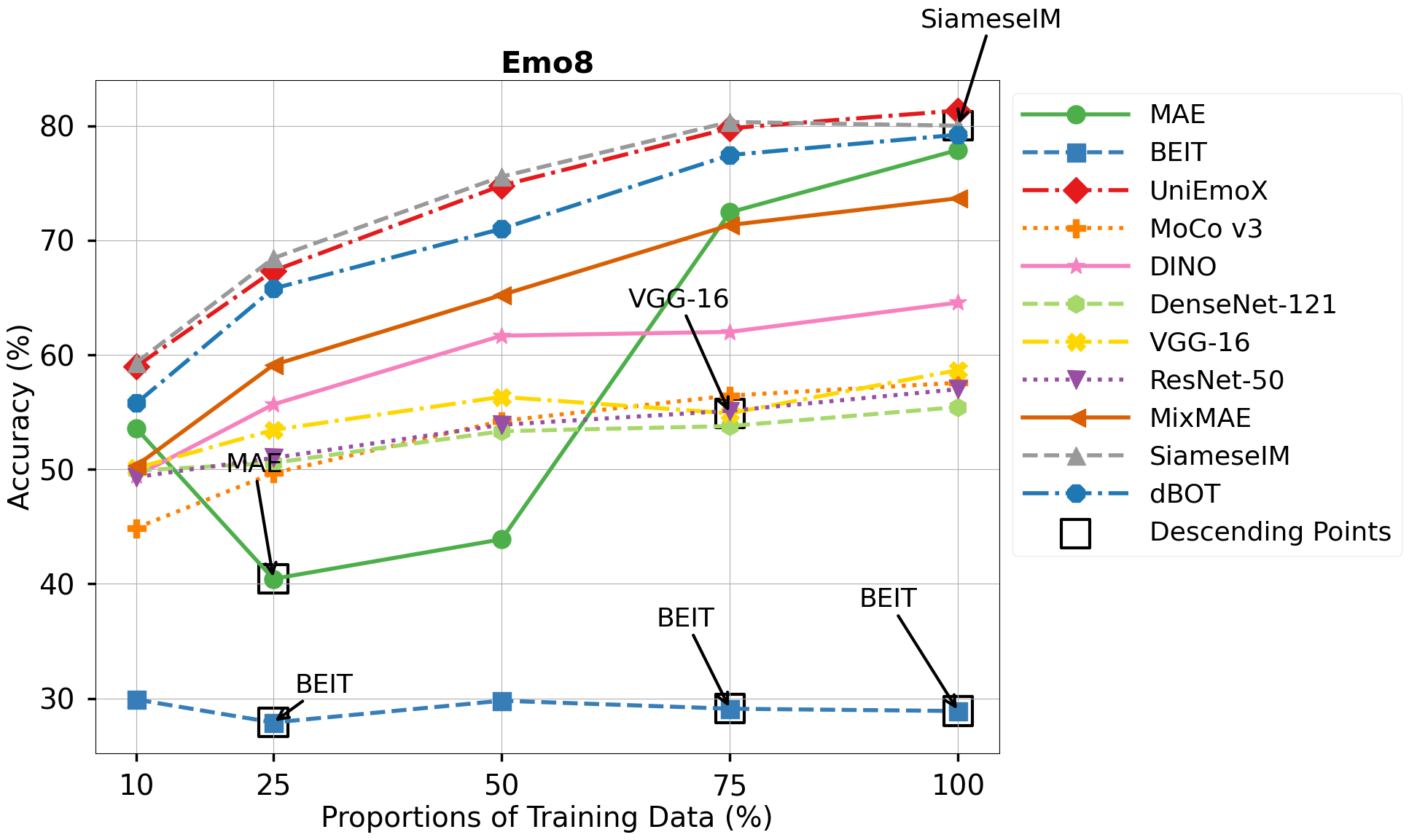} 
  \end{minipage}
  \begin{minipage}[b]{0.329\textwidth}
      \centering
      \includegraphics[width=\linewidth]{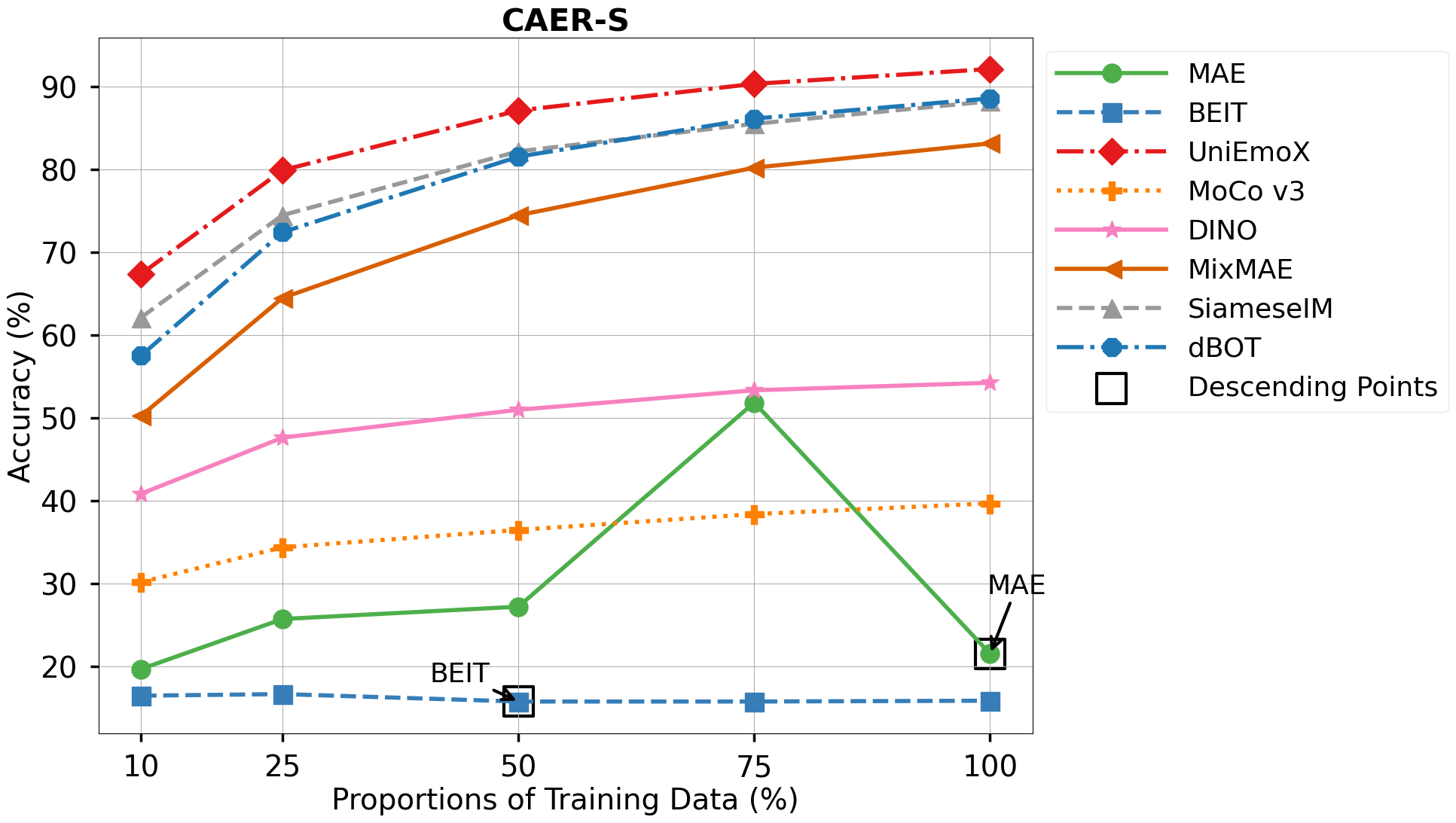} 
  \end{minipage}
  \begin{minipage}[b]{0.329\textwidth}
    \centering
    \includegraphics[width=\linewidth]{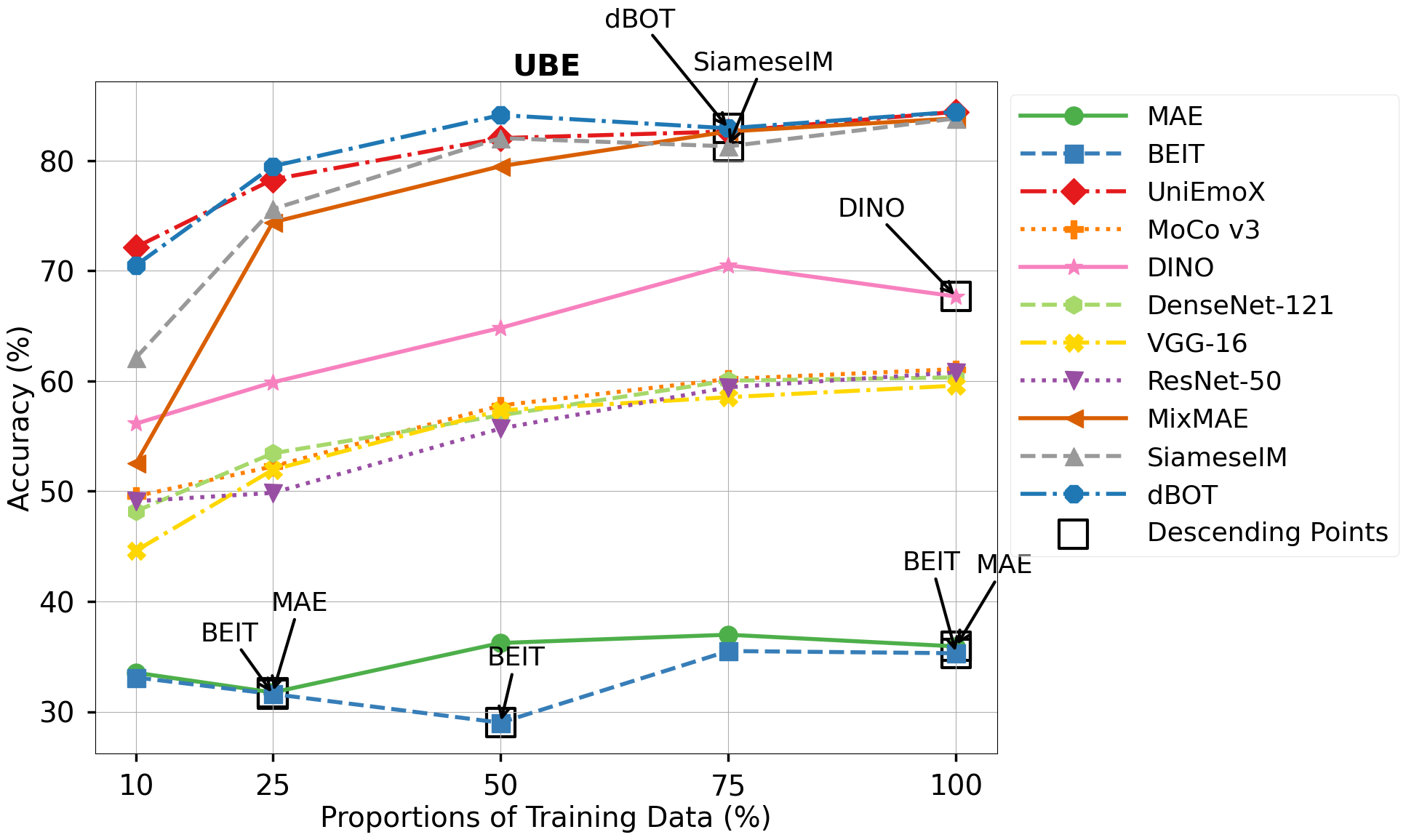} 
  \end{minipage}

  \begin{minipage}[b]{0.329\textwidth}
      \centering
      \includegraphics[width=\linewidth]{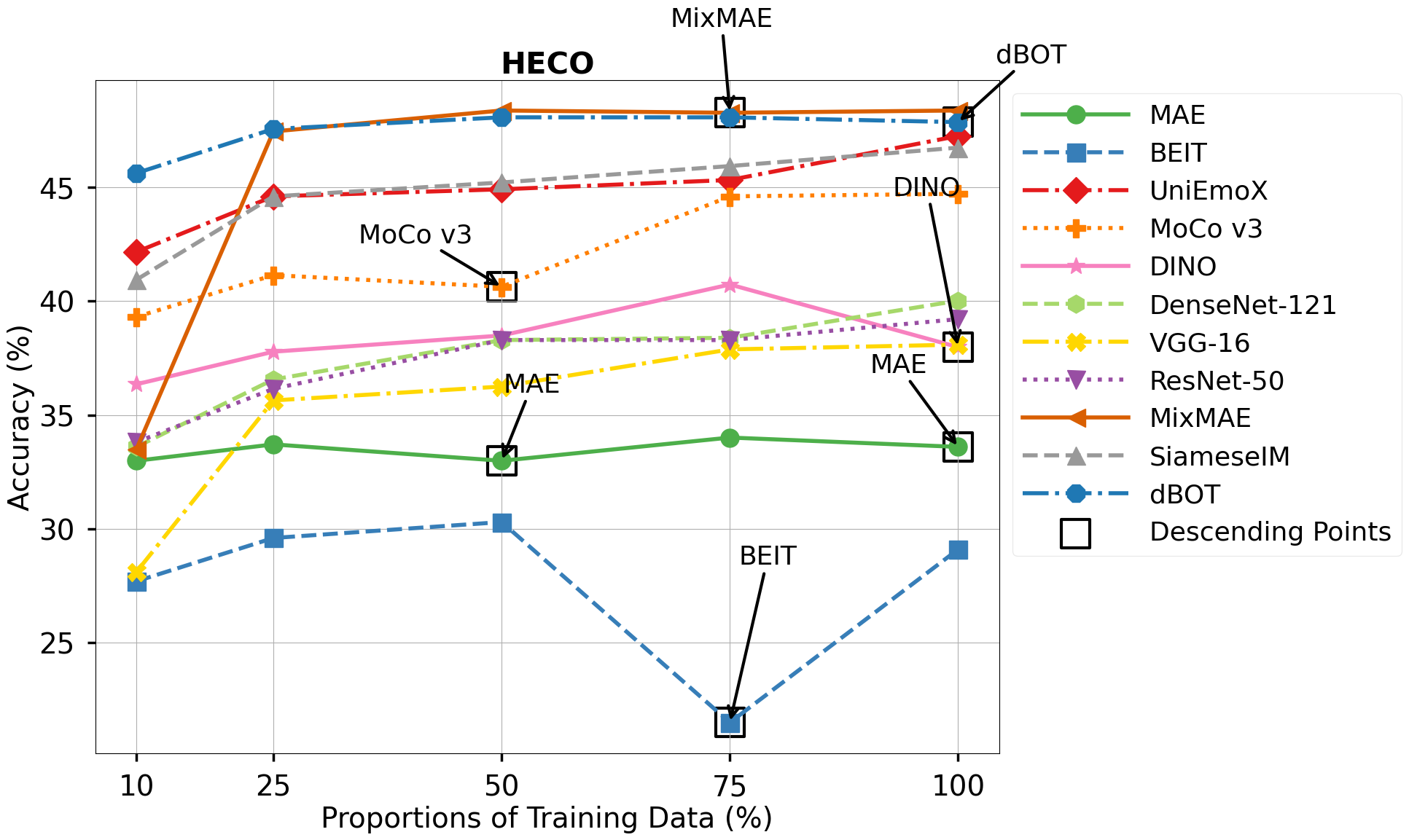} 
  \end{minipage}
  \begin{minipage}[b]{0.329\textwidth}
      \centering
      \includegraphics[width=\linewidth]{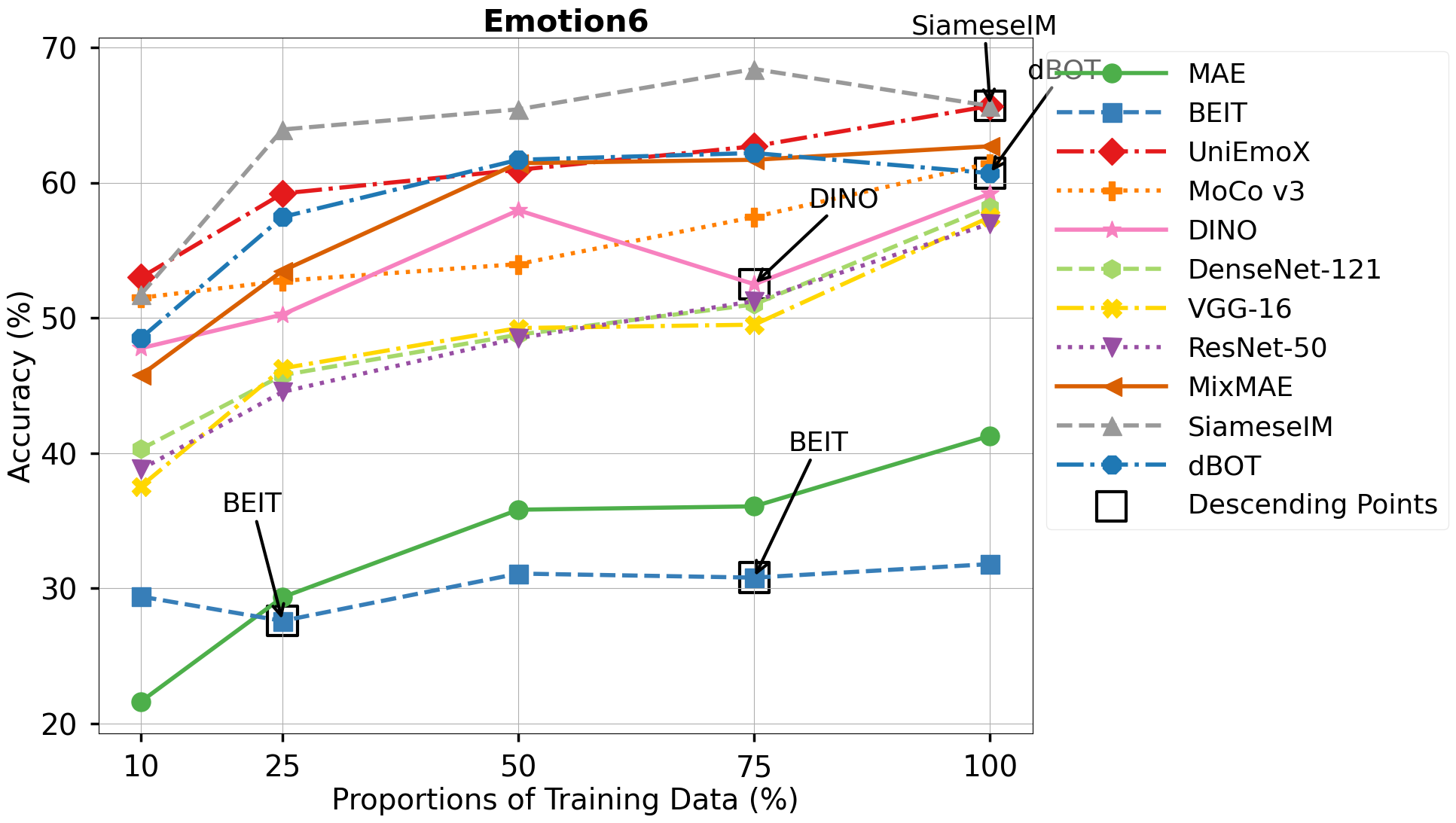} 
  \end{minipage}
  \begin{minipage}[b]{0.329\textwidth}
      \centering
      \includegraphics[width=\linewidth]{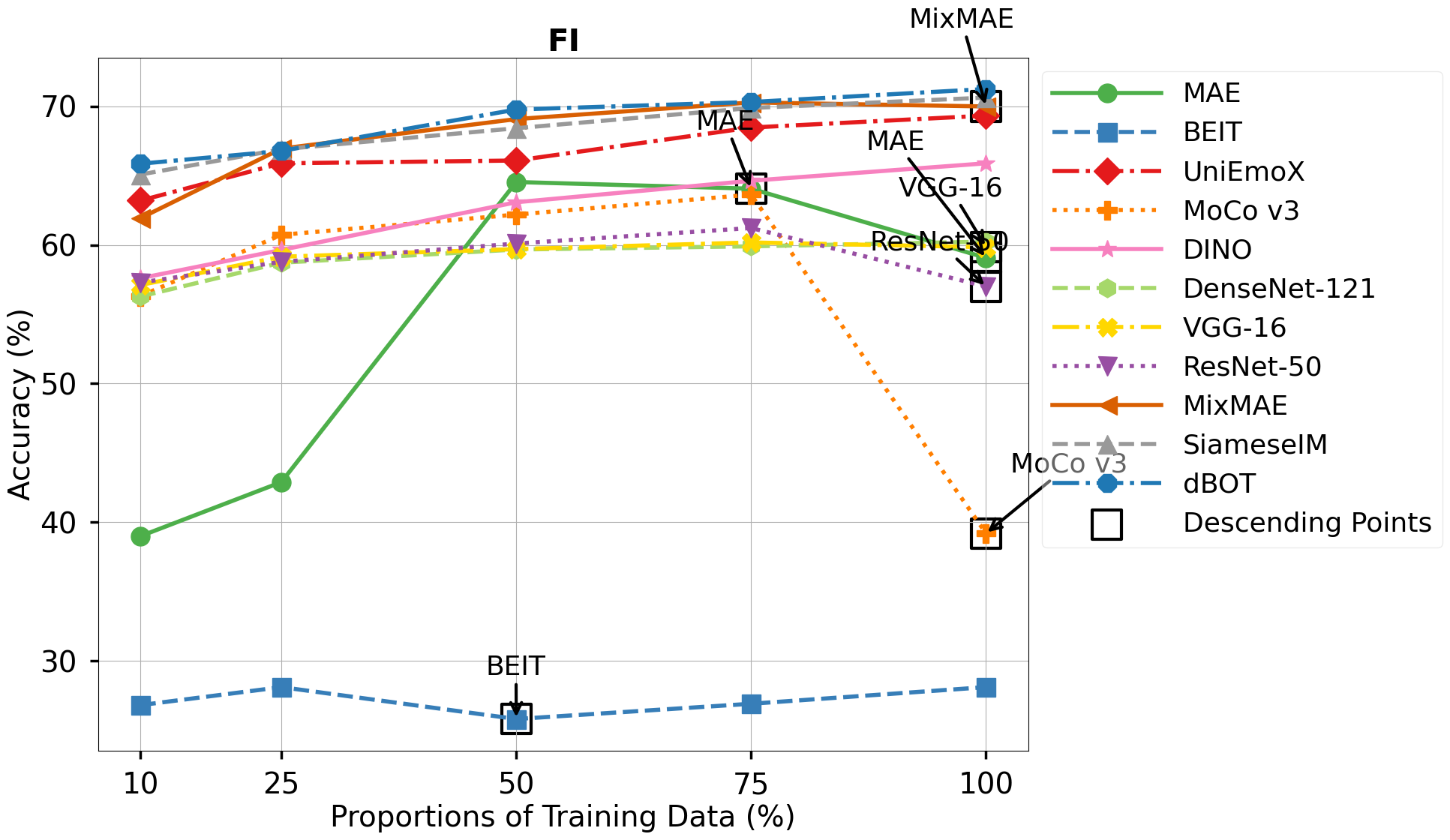} 
  \end{minipage}
  \caption{Evaluating the accuracy of various methods with different proportions of training data on the Emo8, CAER-S, UBE, HECO, Emotion6 and FI datasets respectively.}
  \label{fig:train_data}
  \vspace{-15pt} 
\end{figure*}
\begin{figure}[htbp]
  \begin{center}
  \includegraphics[width=\linewidth]{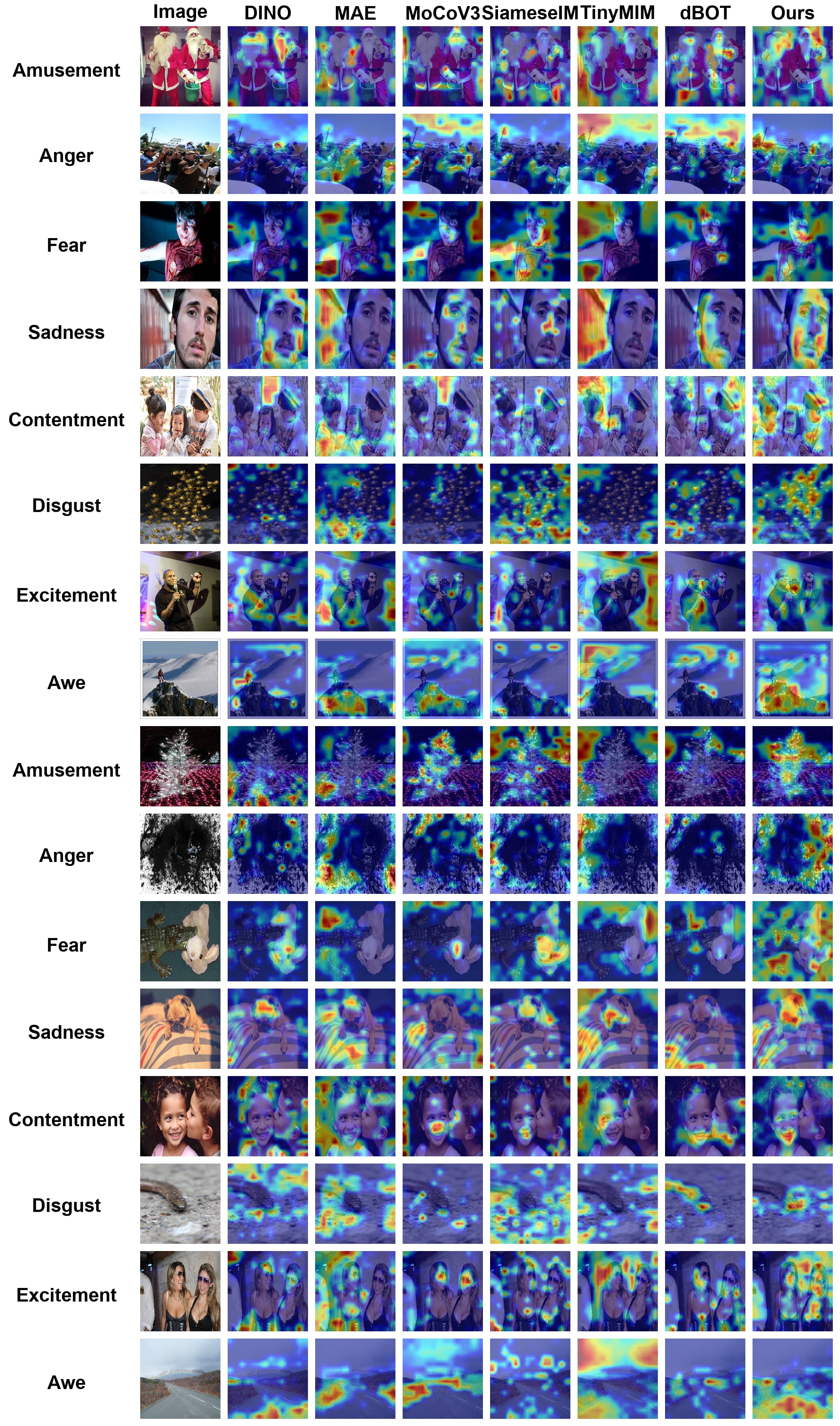}
  \end{center}
   \caption{Comparison of salient regions between UniEmoX and other methods.}
  \vspace{-15pt} 
  \label{fig:mps}
\end{figure}

\begin{figure}[htbp]
  \begin{center}
  \includegraphics[width=\linewidth]{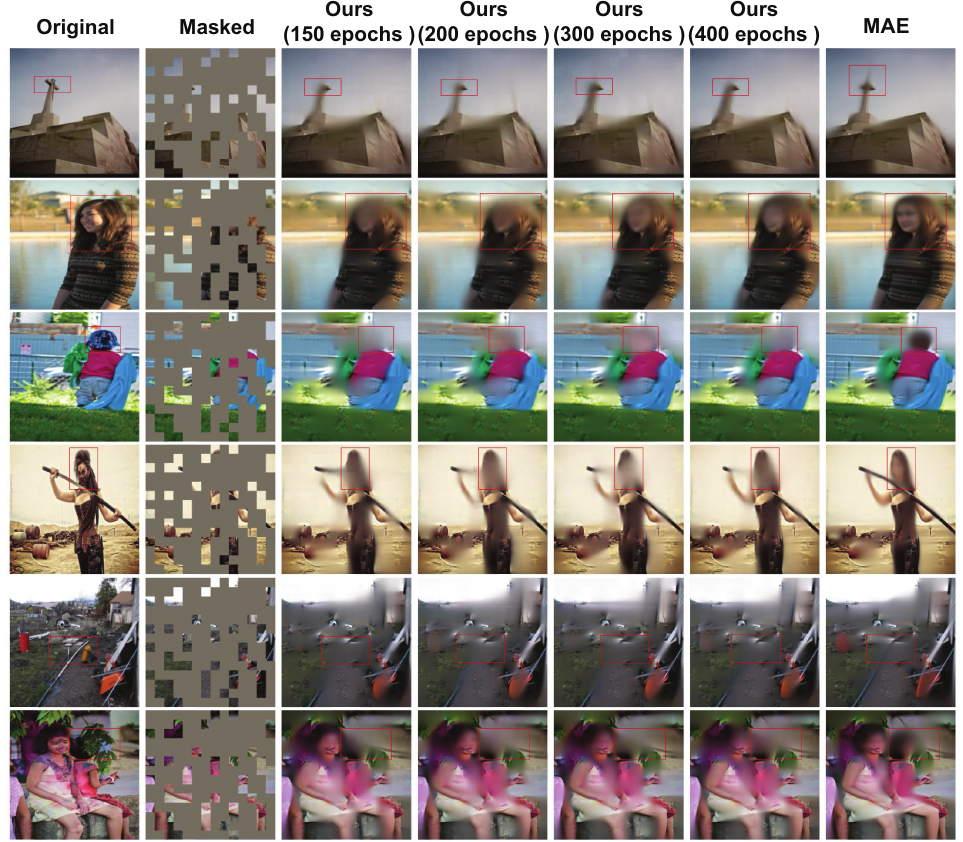}
  \end{center}
  \caption{Comparison of image reconstruction between UniEmoX and MAE.}
  \vspace{-15pt} 
  \label{fig:maskrec}
\end{figure}

\subsection{Comparisons with Previous Results}
We compare our UniEmoX with twenty existing methods, 
including four traditional backbone networks based on supervised pretraining methods (VGG-16~\cite{simonyan2014very}, ResNet-50~\cite{he2016deep}, DenseNet-121~\cite{huang2017densely}, and ViT~\cite{dosovitskiy2020image}), 
ten methods based on self-supervised pretraining methods (MoCoV3~\cite{xinlei2021empirical}, DINO~\cite{caron2021emerging}, BEIT~\cite{bao2021beit}, MAE~\cite{he2022masked}, 
SimMIM~\cite{9880205}, TinyMIM~\cite{10204508}, SiameseIM~\cite{10205066}, MixMAE~\cite{10204021}, dBOT~\cite{liu2024exploring} and CAE~\cite{chen2024context}) 
and six state-of-the-art (SOTA) visual emotion analysis (VAE) methods (SOLVER~\cite{9580604}, Stimuli-aware~\cite{9524517}, MDAN~\cite{9880150}, EAP~\cite{zhang2022visual}, RRLA~\cite{9373919}, and EmotiCon + CCIM~\cite{10636065}). 
BEIT, MAE, SimMIM, SiameseIM, MixMAE, and CAE are self-supervised pretraining methods based on masked image modeling, 
while MoCov3, DINO and dBOT are a pretraining method based on contrastive learning. To ensure fairness and generality, 
pretrained models for the traditional backbone networks are sourced from the official PyTorch repository, 
whereas pretrained models for the self-supervised methods are obtained directly from the original authors.
We fine-tune all pretrained models in our environment. As the source code for previous SOTA methods in VAE is not publicly available, 
their results are cited from the original papers. It is noteworthy that, to validate the effectiveness of the pre-trained model, 
we also train a ViT-L model from scratch using training data from multiple datasets and evaluate it on the corresponding test sets.
The experimental results are shown
in TABLE~\ref{tab:sota}. Based on these results, we can draw
three conclusions:
\begin{enumerate}{}{}
  \item Our UniEmoX outperforms twenty methods across four downstream datasets (UBE, Emotion6, Emo8, CAER-S), 
  including four traditional supervised pre-training methods, ten self-supervised pre-training methods, 
  and six state-of-the-art (SOTA) methods in visual emotion analysis. 
  Furthermore, on two additional downstream datasets (FI and HECO), 
  UniEmoX surpasses the majority of other methods. Overall, UniEmoX achieves the best performance.
  \item We observe that although not all self-supervised methods outperform supervised methods, 
  self-supervised pretraining methods generally achieve superior results. 
  We hypothesize that self-supervised pretraining tasks, designed to learn general data representations, 
  compel the model to learn broader and more generalized features, 
  which may explain their superior performance across different datasets.
  \item We find that both self-supervised and unsupervised pre-training methods, 
  even when based on the ViT architecture, significantly outperform ViT networks trained from scratch. 
  This further validates the importance of large-scale pre-training. 
\end{enumerate}

Furthermore, we aim not only to surpass existing methods in terms of final performance 
but also to develop a 
universal emotional pretraining framework with strong generalization capabilities. 
This framework should learn effectively from smaller proportions of data in downstream emotion recognition tasks. 
To this end, we fine-tune various methods using different proportions of datasets, 
and the experimental results are shown in Fig.~\ref{fig:train_data}. Based on these results, we can draw two conclusions:
\begin{enumerate}{}{}
\item Our UniEmoX demonstrates excellent learning ability with smaller proportions of data in the UBE, Emo8, and CAER-S datasets. 
\item Unlike other methods, UniEmoX consistently demonstrates performance improvements during the fine-tuning phase 
as the increasing proportions of training data, 
without encountering performance bottlenecks (the points indicated by black arrows in 
Fig.~\ref{fig:train_data} represent fluctuations in performance). 
This underscores its exceptional capability for continual learning and generalization. 
\end{enumerate}

\subsection{Ablation Study}
To verify the effectiveness of the various component combinations, we conduct ablation experiments. 
The specific steps are as follows: 
we initialize with a pre-trained model and conduct 200 epochs of multimodal pre-training with a diverse set of components, 
followed by fine-tuning on multiple downstream datasets.
The experimental results are shown in TABLE~\ref{tab:ablation}. Based on these results, we can draw three conclusions:
\begin{enumerate}{}{}
\item By comparing w/o $\mathcal{L}_{1}$ vs. w/ $\mathcal{L}_{1}$, w/o $\mathcal{L}_{2}$ vs. w/ $\mathcal{L}_{2}$, w/o $\mathcal{L}_{3}$ vs. w/ $\mathcal{L}_{3}$, and w/o fusion vs. w/ fusion, 
we systematically evaluate the contributions of Image Reconstruction, Semantic Knowledge, and Psychological Prior 
to the model's performance. Notably, the Semantic-aware Visual Alignment (SVA) Loss $\mathcal{L}_{3}$ plays the most critical 
role within the module combination.

\item Although certain component ($\mathcal{L}_{1}$, $\mathcal{L}_{2}$, and fusion) exhibit performance degradation on specific datasets 
such as CAER-S, UBE, and FI, their overall performance demonstrates an upward trend across 
all datasets on average. This result provides strong evidence supporting the stability of the proposed components.

\end{enumerate}

To assess the impact of masking ratios and various fusion strategies on visual downstream emotion analysis tasks, 
we conduct ablation experiments. The specific steps are as follows: 
the model is first randomly initialized, followed by 150 epochs of pre-training using different 
masking ratios and fusion strategies, and then fine-tuning the pre-trained model on downstream datasets.
The experimental results are shown in TABLE.~\ref{table:ratioMaskResults} and TABLE.~\ref{table:fusionResults}. 
Based on these results, we can draw one conclusion:
\begin{enumerate}{}{}
\item Although MAE research suggests that models pre-trained with a 0.75 masking ratio perform better during fine-tuning, 
we find that models pre-trained with a 0.5 masking ratio learn emotional representations more effectively 
in SVEA datasets. Conversely, models pre-trained with a 0.75 masking ratio 
excel in fine-tuning PVEA datasets.

\item $\Gamma_{1}$ and $\Gamma_{2}$ exhibited superior performance, 
with $\Gamma_{2}$  proving more effective for datasets characterized by significant semantic gaps. 
This is particularly relevant for HECO and FI, which possess relatively large data volumes and 
are not subjected to stringent data quality filtering. 

\item While $\Gamma_{3}$ utilizes additional parameters, 
its performance on downstream datasets is suboptimal. 
Therefore, $\Gamma_{2}$ serves as a robust fusion method, 
more adaptable to datasets with varying qualities, cues, and semantic gaps.

\item Conversely, $\Gamma_{4}$, as a baseline method with minimal parameters, 
underperform across most downstream datasets and is more suitable for addressing smaller semantic gaps. 
More suitable for datasets like CAER-S, where image content is relatively simple and semantic cues are limited.

\item Overall, $\Gamma_{2}$ and $\Gamma_{3}$ outperform $\Gamma_{1}$ and $\Gamma_{4}$, as they incorporate additional parameters that 
enable the learning of more patterns, thereby providing stable performance across various data quality gaps. 
While $\Gamma_{4}$ and $\Gamma_{1}$ perform well on certain datasets, 
they tend to suffer from performance fluctuations when faced with datasets 
characterized by more complex semantic gaps, such as HECO and UBE.

\end{enumerate}

The text descriptions corresponding to emotional images are generated by logically 
linking multiple emotion-related attribute labels using natural language. 
These attributes include brightness, color saturation, scene type, object type, 
facial expressions, action type, and overall emotional expression. Among them, 
brightness, color saturation, scene type, object type, facial expressions, 
and action type are intrinsic image features, present at both the content and structural levels. 
These attributes are predicted by the model rather than manually annotated. 
Therefore, when these text descriptions—constructed through natural language connections—are used 
for UniEmoX pretraining, no additional supervision signals are introduced, 
and no semantic knowledge is leaked. As a result, the entire process can be regarded as fully self-supervised learning.
However, in our previous experiments, the text descriptions included overall emotional expression, 
which may have introduced some degree of semantic information. 
To further assess its impact on model performance, we conducted an ablation experiments. 
The experimental results are shown in TABLE.~\ref{table:emotion_label}. 
Based on these results, we can draw one conclusion:
\begin{enumerate}{}{}
\item The results show that the impact varies across the six datasets, 
with the largest performance difference being approximately two percentage points. 
We hypothesize that the limited impact is due to the relatively small proportion of 
overall emotional expression within the complete natural language descriptions, 
which may not significantly interfere with the model’s learning process.
\end{enumerate}

\subsection{Visualization}
To verify the effectiveness of our UniEmoX more intuitively, 
we perform a visual analysis of the regions the model focuses on in input images by 
generating heatmaps using Grad-CAM\footnote{\href{https://pypi.org/project/pytorch-gradcam/}{https://pypi.org/project/pytorch-gradcam/}}. 
The specific steps are as follows: We download pretrained models for three methods (DINO, MAE, MoCo v3) 
as well as UniEmoX. 
We transfer the parameters of the pretrained models to the corresponding ViT networks and add appropriate 
linear layers based on the number of classification labels. 
We set the last block of ViT as the target layer and ran Grad-CAM to generate the respective feature maps. 
To ensure fairness and generality, all original images are sourced from the EmoSet test set, 
and we randomly select multiple image samples from different labels. 
The visualization results are shown in Fig.~\ref{fig:mps}. Based on these results, we can draw two conclusions:
\begin{enumerate}{}{}
  \item Compared to other methods, UniEmoX focuses on more salient regions. 
  \item Additionally, in terms of judging the emotions elicited 
  by each sample, the salient regions identified by UniEmoX align more closely with human judgment.
\end{enumerate}

Given that our UniEmoX is based on a masked image modeling task, 
we conduct a visual analysis to verify its image reconstruction capabilities. 
The specific steps are as follows: we primarily compare our method with the MAE. 
We use the pretrained MAE model provided by the original authors and our method's pretrained models obtained at 
different pretraining stages. To ensure fairness and generality, all test images are sourced from the EmoSet test set. 
Using the visualization scripts provided by MAE, we visualize the reconstruction capabilities of all pretrained models 
on the test samples, as shown in Fig.~\ref{fig:maskrec}. Based on these results, we can draw two conclusions:
\begin{enumerate}{}{}
  \item The reconstruction performance of our pretrained model appears inferior to that of MAE. 
  This is because MAE's training process utilizes a larger dataset and focused on optimizing image reconstruction capability. 
  In contrast, our UniEmoX uses a smaller dataset and has multiple optimization objectives, 
  resulting in weaker reconstruction performance. 
  \item We also observe that MAE's reconstruction results exhibit certain 
  hallucination phenomena. As shown in the red box in Fig.~\ref{fig:maskrec}, 
  MAE generates some unrealistic regions during image reconstruction. 
  Therefore, compared to our UniEmoX, which solely optimizes image reconstruction capability, 
  we believe that MAE might have overfitting issues in emotion recognition tasks.
\end{enumerate}
\begin{figure}[htbp]
  \begin{center}
  \includegraphics[width=\linewidth]{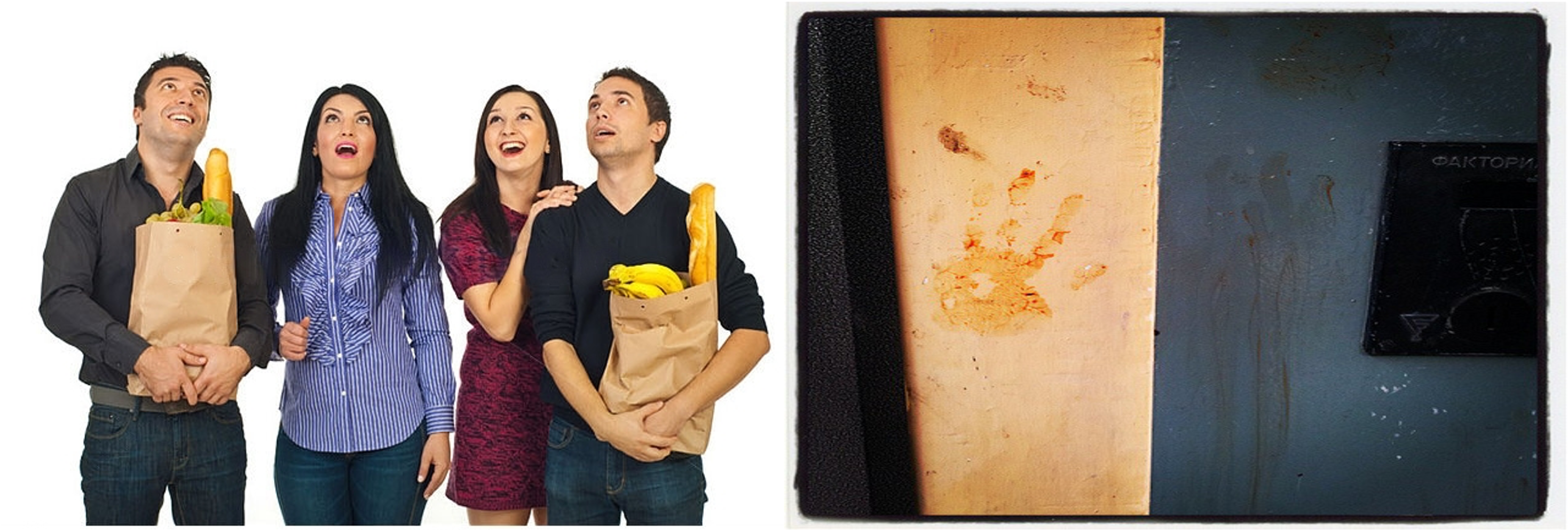}
  \end{center}
  \caption{The left image is taken from the UBE test set and is labeled with the ground truth 'Surprise,' while the right image is from the FI test set and is labeled with the ground truth 'Fear.'}
  \label{fig:modelsample}
  \vspace{-15pt} 
\end{figure}
\begin{table*}
    \centering
    \caption{The evaluation of various multimodal large models in simplified dialogue scenarios examines emotion classification results, causal analysis, and the performance of our proposed UniEmoX model. The table displays the results related to the left image in \textsc{Fig.}~\ref{fig:modelsample}.}
    \label{tab:byube}
    \resizebox{1.0\textwidth}{!}{
    \begin{tabular}{lccc} 
        \hline \toprule [0.3 pt]
        \textbf{Model} & \textbf{Emotion} & \textbf{Description} \\  
        \midrule  
        \href{https://yiyan.baidu.com/}{ERNIE Bot}\textsubscript{\textit{Baidu}} 
        & Surprise & \parbox{15cm}{The facial expressions of the first and last males, as well as the second female, all suggest surprise or amazement. Their wide eyes, open mouths, and turned heads indicate that they are reacting to something unexpected or noteworthy. The third female's raised eyebrows and upward gaze also hint at curiosity or interest in whatever it is that has captured their attention. The overall atmosphere of the scene, with the figures facing in the same direction and seemingly in response to an external stimulus, further reinforces the sense of surprise and wonder. None of the other listed emotions ('sadness', 'fear', 'joy', 'love', 'anger') are as clearly represented in the image as surprise is.} \\ 
        \href{https://xinghuo.xfyun.cn/desk}{IFlytek Spark}\textsubscript{\textit{IFlytek}} 
        & Surprise &\parbox{15cm}{This is inferred from the wide eyes and open mouths of each person, which are common facial expressions associated with surprise. The upward gaze also suggests a heightened sense of curiosity or wonder. While there is no overt sign of joy, love, fear, or anger, the overall atmosphere of the image does not align with these emotions either.}\\
        \href{https://chatglm.cn/main/alltoolsdetail?lang=zh}{ChatGLM} \textsubscript{\textit{Tsinghua}} 
        & Joy &\parbox{15cm}{The expressions on their faces are bright and cheerful, with wide smiles and eyes looking upwards as if they are excited about something. There is no indication of sadness, fear, love, anger, or surprise in their expressions.}   \\
        \href{https://tongyi.aliyun.com/}{Qwen}\textsubscript{\textit{Alibaba}} 
        & Surprise & \parbox{15cm}{The group of people holding grocery bags and looking up with a surprised expression suggests that they are reacting to something unexpected or interesting above them. The other emotions listed - sadness, fear, joy, love, and anger - are not as prominent in the scene, as their facial expressions and body language indicate surprise rather than any of those particular emotions.}\\
        \href{https://ying.baichuan-ai.com/chat?from=home}{Baichuan}\textsubscript{\textit{Baichuan AI}} 
        & Joy & \parbox{15cm}{This conclusion is based on the visual cues present in the individuals' facial expressions and body language. Each person is smiling broadly, which is a common indicator of happiness or joy. Their eyes appear to be crinkled in a way that suggests genuine amusement, and their open mouths could indicate laughter. The context of holding grocery bags does not inherently suggest any negative emotions such as sadness, fear, anger, or surprise. Therefore, the most intense emotional response conveyed by the image is joy.}  \\ 
        \href{https://chat.sensetime.com/}{SenseChat}\textsubscript{\textit{SenseTime}} 
        & Joy & \parbox{15cm}{Reasoning: 
        1. Facial Expressions: All four individuals in the image are smiling and have their mouths open, which is a common expression associated with happiness or joy. 
        2. Body Language: Their body language appears relaxed and content. The man on the left has his hands clasped together while holding a bag, suggesting a casual and happy demeanor. The woman next to him also seems cheerful as she looks upwards with an open-mouthed smile. The other two individuals exhibit similar expressions of delight. 
        These visual cues collectively indicate that the predominant emotional response elicited by this image is one of joy.} \\ 
        \href{https://chatgpt.com/}{GPT-4o}\textsubscript{\textit{OpenAI}} 
        & Surprise & \parbox{15cm}{The reason for this is that the people in the photo are looking upward with wide eyes and open mouths, expressions typically associated with surprise. Their body language, including the way they are holding their groceries and the way they are leaning slightly backward, further suggests that they are reacting to something unexpected above them.} \\ 
        UniEmoX (ours) & Surprise & \\ 
        \hline \toprule [0.3 pt]
    \end{tabular}
    }
    \vspace{-10pt} 
\end{table*}
\begin{table*}
    \centering
    \caption{The evaluation of various multimodal large models in simplified dialogue scenarios examines emotion classification results, causal analysis, and the performance of our proposed UniEmoX model. The table displays the results related to the right image in \textsc{Fig.}~\ref{fig:modelsample}.}
    \label{tab:byfi}
    \resizebox{1.0\textwidth}{!}{
    \begin{tabular}{lccc} 
        \toprule  
        \textbf{Model} & \textbf{Emotion} & \textbf{Description} \\  
        \hline \toprule [0.3 pt]
        \href{https://yiyan.baidu.com/}{ERNIE Bot}\textsubscript{\textit{Baidu}} 
        & Fear & \parbox{15cm}{The reason is that the image depicts a scene that is both mysterious and potentially threatening. The yellow wall with a handprint-shaped stain could suggest that someone or something has been there recently, leaving behind a trace of their presence. The black object with the unusual "GAKTOP" (or possibly "PAKTOPV" as mentioned in the additional information) label on the blue wall adds to the sense of unease, as it could represent an unknown device or mechanism. The overall atmosphere of the image, with its worn-down and abandoned appearance, creates a sense of isolation and potential danger. 
        The handprint could be interpreted as a sign of someone who was in distress or trying to escape, which could evoke feelings of fear for the person's safety. The unknown nature of the black object and its label also contribute to the sense of unease and potential threat. The other emotions listed, such as amusement, anger, awe, contentment, disgust, excitement, and sadness, do not seem to fit the emotional tone conveyed by the image as strongly as fear does.} \\ 
        \href{https://xinghuo.xfyun.cn/desk}{IFlytek Spark}\textsubscript{\textit{IFlytek}} 
        & Disgust &\parbox{15cm}{}\\
        \href{https://chatglm.cn/main/alltoolsdetail?lang=zh}{ChatGLM}\textsubscript{\textit{Tsinghua}}
        & Disgust  &\parbox{15cm}{The presence of a handprint on what appears to be a door, possibly indicating a recent and unsettling event, contributes to this feeling. Additionally, the door’s grimy condition suggests neglect or decay, which can also lead to feelings of disgust.}   \\
        \href{https://tongyi.aliyun.com/}{Qwen}\textsubscript{\textit{Alibaba}} 
        & Disgust & \parbox{15cm}{The dirty handprint on the wall and the overall grungy appearance of the scene might make people feel uncomfortable due to the uncleanliness and lack of maintenance. The other emotions listed - amusement, anger, awe, contentment, excitement, and sadness - are not as prominently displayed in the image, as the focus is on the dirty handprint and its surrounding area.}\\
        \href{https://ying.baichuan-ai.com/chat?from=home}{Baichuan}\textsubscript{\textit{Baichuan AI}} 
        & Disgust & \parbox{15cm}{This is because the sight of a handprint made in what appears to be fecal matter is inherently repulsive. The act of someone placing their hand in such material and then onto a wall is considered unhygienic and grossly inappropriate, which would likely elicit a strong visceral reaction of disgust from the viewer.}  \\ 
        \href{https://chat.sensetime.com/}{SenseChat}\textsubscript{\textit{SenseTime}}
        & Disgust & \parbox{15cm}{Reasoning: 1. Visual Elements: The handprint on the wall appears to be made with a substance that looks like blood or some other bodily fluid, which is inherently unsettling and disgusting. 
        2. Contextual Clues: The presence of what seems to be a fire alarm box next to it suggests an emergency situation, possibly involving injury or violence, further contributing to feelings of disgust.}\\
        \href{https://chatgpt.com/}{GPT-4o}\textsubscript{\textit{OpenAI}} 
        & Fear & \parbox{15cm}{The reason for this is the presence of a handprint that appears to be smeared with a red substance, resembling blood, on a surface in what seems to be an industrial or abandoned setting. This handprint suggests a sense of urgency, distress, or a violent event. The context of the handprint and the overall dark and gritty atmosphere contribute to a feeling of fear and unease.} \\ 
        UniEmoX (ours) & Fear & \\ 
        \hline \toprule [0.3 pt]
    \end{tabular}
    }
   
    \vspace{-10pt} 
\end{table*}

\subsection{Comparative Analysis of Emotion Understanding Capabilities in Multimodal Large Models}
Many multimodal large models, such as \href{https://chatgpt.com/}{GPT-4o}, \href{https://yiyan.baidu.com/}{ERNIE Bot3.5}, 
\href{https://xinghuo.xfyun.cn/desk}{iFlytek Spark}, \href{https://chatglm.cn/main/alltoolsdetail}{ChatGLM}, 
\href{https://tongyi.aliyun.com/}{Qwen2.5}, \href{https://ying.baichuan-ai.com/chat?from=home}{Baichuan4} 
and \href{https://chat.sensetime.com/}{SenseChat} are currently deployed in practical applications. 
These models accurately interpret user intentions in complex conversational contexts by integrating data from various 
modalities. To assess their performance in emotional semantic understanding, we design a straightforward 
dialogue scenario. We evaluate their accuracy in emotion recognition and depth of emotional semantic understanding 
through image-text question-and-answer interactions. Specifically, we ask: “Which emotion does this image most strongly 
evoke? Please select the most intense emotional response from the following list and explain your choice: 
['sadness', 'fear', 'excitement', 'disgust', 'contentment', 'awe', 'anger', 'amusement']." We randomly select images 
from the FI test set, input them into the models, and record the responses. To ensure result generalizability, 
we repeat the test with the UBE test set. 
Due to space constraints, we present analysis results for only two examples per model. Detailed results for additional examples are provided in Appendix Section IV. 
By analyzing the performance of these multimodal models alongside our UniEmoX, we can draw several key conclusions:
\begin{enumerate}{}{}
  \item Multimodal large models demonstrate exceptional accuracy in emotion recognition and depth of emotional semantic understanding on most test samples. 
  However, they still have limitations. 
  For instance, some models demonstrate limited depth in emotional understanding.
  As shown in the left image of Fig.~\ref{fig:modelsample} and TABLE~\ref{tab:byube},
  most viewers might initially interpret the character's 
  facial expression as indicating happiness. Consequently, models such as ChatGLM, Baichuan, and SenseChat classify it as 'Joy'. 
  However, this interpretation is superficial. A more nuanced analysis reveals that the character's upward gaze suggests a reaction to 
  something unexpected above. Therefore, the most accurate emotional label for this image is 'Surprise'. 
  Notably, ERNIE Bot, iFlytek Spark, Qwen, and GPT-4o correctly identified this emotion.
  \item  As shown in the right image of Fig.~\ref{fig:modelsample} and TABLE~\ref{tab:byfi},
  the initial impression might be that the chaotic environment evokes a sense of 'disgust', 
  which aligns with the assessments of iFlytek Spark, ChatGLM, Qwen, Baichuan, and SenseChat. 
  However, the presence of a crucial bloodstain and the eerie ambiance of the environment more 
  accurately elicit a feeling of 'fear'. ERNIE Bot and GPT-4o correctly identified this emotion.
  \item The two cases discussed above reveal that some multimodal large models tend to remain superficial in their emotional 
  semantic understanding, lacking deeper analysis. 
  Additional test results are provided in Appendix Section IV due to space constraints.
  In contrast, despite having significantly fewer parameters and 
  lower training costs compared to many large models, our UniEmoX accurately identifies the emotions in these samples. 
  This suggests that our model offers a distinct advantage in emotion recognition tasks.
\end{enumerate}

\section{Conclusion}\label{conclusion} 
We introduce UniEmoX, a large-scale pretraining framework guided by cross-modal semantics, 
designed to address the poor generalization of existing visual emotion analysis methods, 
which is caused by the ambiguity of emotional perception and the diversity of data scenes.
Inspired by psychological research emphasizing the inseparability of the emotional 
exploration process from the interaction between individuals and their environment, UniEmoX integrates 
scene-centric and person-centric low-level image spatial structural information, aiming to derive higher-level 
latent representations. By exploiting the similarity between paired and unpaired image-text samples, 
UniEmoX distills rich semantic knowledge from the CLIP model to enhance emotional embedding representations 
in a more effective manner. We have developed a visual emotional dataset titled Emo8. 
Emo8 samples cover a range of domains, including advertising images, natural scenes, portraits, and cartoons. 
Comprehensive experiments conducted on six benchmark datasets across two downstream tasks validate the effectiveness of UniEmoX.

\section{Limitation}\label{limitation} 
In constructing the emotional pretraining framework for visual perception, 
guided by three types of semantic information (namely, structural information in image space, 
textual-image semantic information, and psychological theory prior knowledge), 
we successfully develop an emotional pretraining framework tailored 
for large-scale complex environments. While the experiments validate the progress of 
our UniEmoX, opportunities for enhancement remain. From the perspective of information granularity, 
the three types of semantic information in our UniEmoX are relatively coarse. 
Future efforts will focus on extracting finer-grained emotional-relevant regions 
from entire scene information, such as expressions, gazes, gaits, etc., 
to enhance the emotional expression capability of pretraining models. 
Additionally, future efforts will focus on constructing larger-scale and 
more diverse emotional datasets with richer emotional annotations.

\section{Acknowledgments}\label{acknowledgments} 
This work was supported by Special Project of the National Natural Science Foundation of China (62441614), 
Anhui Province Key R\&D Program (202304a05020068) and General Programmer of the National Natural Science 
Foundation of China (62376084).
\bibliographystyle{IEEEtran}
\bibliography{IEEEabrv, references}

\begin{thebibliography}{10}
\providecommand{\url}[1]{#1}
\csname url@samestyle\endcsname
\providecommand{\newblock}{\relax}
\providecommand{\bibinfo}[2]{#2}
\providecommand{\BIBentrySTDinterwordspacing}{\spaceskip=0pt\relax}
\providecommand{\BIBentryALTinterwordstretchfactor}{4}
\providecommand{\BIBentryALTinterwordspacing}{\spaceskip=\fontdimen2\font plus
\BIBentryALTinterwordstretchfactor\fontdimen3\font minus \fontdimen4\font\relax}
\providecommand{\BIBforeignlanguage}[2]{{%
\expandafter\ifx\csname l@#1\endcsname\relax
\typeout{** WARNING: IEEEtran.bst: No hyphenation pattern has been}%
\typeout{** loaded for the language `#1'. Using the pattern for}%
\typeout{** the default language instead.}%
\else
\language=\csname l@#1\endcsname
\fi
#2}}
\providecommand{\BIBdecl}{\relax}
\BIBdecl

\bibitem{4523967}
Z.~Zeng, J.~Tu, B.~M. Pianfetti, and T.~S. Huang, ``Audio–visual affective expression recognition through multistream fused {HMM},'' \emph{{IEEE} Trans. Multimedia}, vol.~10, no.~4, pp. 570--577, Jun. 2008.

\bibitem{8052551}
F.~Chen, R.~Ji, J.~Su, D.~Cao, and Y.~Gao, ``Predicting microblog sentiments via weakly supervised multimodal deep learning,'' \emph{{IEEE} Trans. Multimedia}, vol.~20, no.~4, pp. 997--1007, Apr. 2018.

\bibitem{6362231}
M.~Tkalcic, A.~Odic, A.~Kosir, and J.~Tasic, ``Affective labeling in a content-based recommender system for images,'' \emph{{IEEE} Trans. Multimedia}, vol.~15, no.~2, pp. 391--400, Feb. 2013.

\bibitem{9524517}
J.~Yang, J.~Li, X.~Wang, Y.~Ding, and X.~Gao, ``Stimuli-aware visual emotion analysis,'' \emph{{IEEE} Trans. Image Process.}, vol.~30, pp. 7432--7445, 2021.

\bibitem{9580604}
J.~Yang, X.~Gao, L.~Li, X.~Wang, and J.~Ding, ``{SOLVER}: Scene-object interrelated visual emotion reasoning network,'' \emph{{IEEE} Trans. Image Process.}, vol.~30, pp. 8686--8701, 2021.

\bibitem{9846869}
J.~Yang, J.~Li, L.~Li, X.~Wang, Y.~Ding, and X.~Gao, ``Seeking subjectivity in visual emotion distribution learning,'' \emph{{IEEE} Trans. Image Process.}, vol.~31, pp. 5189--5202, 2022.

\bibitem{kosti2019context}
R.~Kosti, J.~M. Alvarez, A.~Recasens, and A.~Lapedriza, ``Context based emotion recognition using {EMOTIC} dataset,'' \emph{{IEEE} Trans. Pattern Anal. Mach. Intell.}, vol.~42, no.~11, pp. 2755--2766, Nov. 2020.

\bibitem{mittal2020emoticon}
T.~Mittal, P.~Guhan, U.~Bhattacharya, R.~Chandra, A.~Bera, and D.~Manocha, ``Emoti{C}on: Context-aware multimodal emotion recognition using frege’s principle,'' in \emph{Proc. {IEEE} Conf. Comput. Vis. Pattern Recognit. (CVPR)}, Jun. 2020, pp. 14\,222--14\,231.

\bibitem{brainerd2018emotional}
C.~Brainerd, ``The emotional-ambiguity hypothesis: A large-scale test,'' \emph{Psychol. Sci.}, vol.~29, no.~10, pp. 1706--1715, 2018.

\bibitem{panda2018contemplating}
R.~Panda, J.~Zhang, H.~Li, J.-Y. Lee, X.~Lu, and A.~K. Roy-Chowdhury, ``Contemplating visual emotions: Understanding and overcoming dataset bias,'' in \emph{Proc. Eur. Conf. Comput. Vis. (ECCV)}, Sep. 2018, pp. 579--595.

\bibitem{he2022masked}
K.~He, X.~Chen, S.~Xie, Y.~Li, P.~Dollár, and R.~Girshick, ``Masked autoencoders are scalable vision learners,'' in \emph{Proc. {IEEE} Conf. Comput. Vis. Pattern Recognit. (CVPR)}, Jun. 2022, pp. 15\,979--15\,988.

\bibitem{radford2021learning}
A.~Radford, J.~W. Kim, C.~Hallacy, A.~Ramesh, G.~Goh, S.~Agarwal, G.~Sastry, A.~Askell, P.~Mishkin, J.~Clark, G.~Krueger, and I.~Sutskever, ``Learning transferable visual models from natural language supervision,'' in \emph{Proc. 38th Int. Conf. Mach. Learn. (ICML)}, Jul. 2021, pp. 8748--8763.

\bibitem{deng2009imagenet}
J.~Deng, W.~Dong, R.~Socher, L.-J. Li, K.~Li, and L.~Fei-Fei, ``Image{N}et: A large-scale hierarchical image database,'' in \emph{Proc. {IEEE} Conf. Comput. Vis. Pattern Recognit. (CVPR)}, Jun. 2009, pp. 248--255.

\bibitem{erhan2010does}
D.~Erhan, A.~Courville, Y.~Bengio, and P.~Vincent, ``Why does unsupervised pre-training help deep learning?'' in \emph{Proc. 13th Int. Conf. Artif. Intell. Statist.}, May 2010, pp. 201--208.

\bibitem{gui2024survey}
J.~Gui, T.~Chen, J.~Zhang, Q.~Cao, Z.~Sun, H.~Luo, and D.~Tao, ``A survey on self-supervised learning: Algorithms, applications, and future trends,'' \emph{{IEEE} Trans. Pattern Anal. Mach. Intell.}, pp. 1--20, early access, Jun. 17, 2024, doi:{\color{blue}\href{http://dx.doi.org/10.1109/TPAMI.2024.3415112}{10.1109/TPAMI.2024.3415112}}.

\bibitem{gidaris2018unsupervised}
S.~Gidaris, P.~Singh, and N.~Komodakis, ``Unsupervised representation learning by predicting image rotations,'' 2018, \textit{arXiv:1803.07728}.

\bibitem{larsson2017colorization}
G.~Larsson, M.~Maire, and G.~Shakhnarovich, ``Colorization as a proxy task for visual understanding,'' in \emph{Proc. {IEEE} Conf. Comput. Vis. Pattern Recognit. (CVPR)}, Jul. 2017, pp. 840--849.

\bibitem{noroozi2016unsupervised}
M.~Noroozi and P.~Favaro, ``Unsupervised learning of visual representations by solving jigsaw puzzles,'' in \emph{Proc. Eur. Conf. Comput. Vis. (ECCV)}, Oct. 2016, pp. 69--84.

\bibitem{barrett2011context}
L.~F. Barrett, B.~Mesquita, and M.~Gendron, ``Context in emotion perception,'' \emph{Curr. Dir. Psychol. Sci.}, vol.~20, no.~5, pp. 286--290, 2011.

\bibitem{9472932}
S.~Zhao, X.~Yao, J.~Yang, G.~Jia, G.~Ding, T.-S. Chua, B.~W. Schuller, and K.~Keutzer, ``Affective image content analysis: Two decades review and new perspectives,'' \emph{{IEEE} Trans. Pattern Anal. Mach. Intell.}, vol.~44, no.~10, pp. 6729--6751, Oct. 2022.

\bibitem{9591550}
S.~Zhao, G.~Jia, J.~Yang, G.~Ding, and K.~Keutzer, ``Emotion recognition from multiple modalities: Fundamentals and methodologies,'' \emph{{IEEE} Signal Process. Mag.}, vol.~38, no.~6, pp. 59--73, Nov. 2021.

\bibitem{you2015robust}
Q.~You, J.~Luo, H.~Jin, and J.~Yang, ``Robust image sentiment analysis using progressively trained and domain transferred deep networks,'' in \emph{Proc. {AAAI} Conf. Artif. Intell. (AAAI)}, vol.~29, no.~1, Feb. 2015.

\bibitem{rao2020learning}
T.~Rao, X.~Li, and M.~Xu, ``Learning multi-level deep representations for image emotion classification,'' \emph{Neural Process. Lett.}, vol.~51, pp. 2043--2061, 2020.

\bibitem{9880150}
L.~Xu, Z.~Wang, B.~Wu, and S.~Lui, ``{MDAN}: Multi-level dependent attention network for visual emotion analysis,'' in \emph{Proc. {IEEE} Conf. Comput. Vis. Pattern Recognit. (CVPR)}, Jun. 2022, pp. 9469--9478.

\bibitem{chen2023ast}
C.~Chen, X.~Sun, Z.~Tu, and M.~Wang, ``{AST-GCN}: Augmented spatial temporal graph convolutional neural network for gait emotion recognition,'' \emph{{IEEE} Trans. Circuits Syst. Video Technol.}, vol.~34, no.~6, pp. 4581--4595, Jun. 2024.

\bibitem{chen2023sta}
C.~Chen and X.~Sun, ``{STA-GCN}: Spatial temporal adaptive graph convolutional network for gait emotion recognition,'' in \emph{Proc. {IEEE} Int. Conf. Multimedia Expo (ICME)}, Jul. 2023, pp. 1385--1390.

\bibitem{tawari2013face}
A.~Tawari and M.~M. Trivedi, ``Face expression recognition by cross modal data association,'' \emph{{IEEE} Trans. Multimedia}, vol.~15, no.~7, pp. 1543--1552, Nov. 2013.

\bibitem{8576656}
Y.~Li, J.~Zeng, S.~Shan, and X.~Chen, ``Occlusion aware facial expression recognition using cnn with attention mechanism,'' \emph{{IEEE} Trans. Image Process.}, vol.~28, no.~5, pp. 2439--2450, May 2019.

\bibitem{WANG2022103679}
Z.~Wang, L.~Lao, X.~Zhang, Y.~Li, T.~Zhang, and Z.~Cui, ``Context-dependent emotion recognition,'' \emph{J. Vis. Commun. Image Represent.}, vol.~89, p. 103679, 2022.

\bibitem{9373919}
W.~Li, X.~Dong, and Y.~Wang, ``Human emotion recognition with relational region-level analysis,'' \emph{{IEEE} Trans. Affect. Comput.}, vol.~14, no.~1, pp. 650--663, Jan./Mar. 2023.

\bibitem{10636065}
D.~Yang, K.~Yang, H.~Kuang, Z.~Chen, Y.~Wang, and L.~Zhang, ``Towards context-aware emotion recognition debiasing from a causal demystification perspective via de-confounded training,'' \emph{{IEEE} Trans. Pattern Anal. Mach. Intell.}, pp. 1--18, early access, Aug. 13, 2024, doi:{\color{blue}\href{http://dx.doi.org/10.1109/TPAMI.2024.3443129}{10.1109/TPAMI.2024.3443129}}.

\bibitem{hervella2024multi}
{\'A}.~S. Hervella, J.~Rouco, J.~Novo, and M.~Ortega, ``Multi-adaptive optimization for multi-task learning with deep neural networks,'' \emph{Neural Netw.}, vol. 170, pp. 254--265, 2024.

\bibitem{larsson2017learningrepresentationsautomaticcolorization}
G.~Larsson, M.~Maire, and G.~Shakhnarovich, ``Learning representations for automatic colorization,'' in \emph{Proc. Eur. Conf. Comput. Vis. (ECCV)}, Oct. 2016, pp. 577--593.

\bibitem{xinlei2021empirical}
X.~Chen, S.~Xie, and K.~He, ``An empirical study of training self-supervised vision transformers,'' in \emph{Proc. {IEEE/CVF} Int. Conf. Comput. Vis. (ICCV)}, Oct. 2021, pp. 9620--9629.

\bibitem{caron2021emerging}
M.~Caron, H.~Touvron, I.~Misra, H.~Jegou, J.~Mairal, P.~Bojanowski, and A.~Joulin, ``Emerging properties in self-supervised vision transformers,'' in \emph{Proc. {IEEE/CVF} Int. Conf. Comput. Vis. (ICCV)}, Oct. 2021, pp. 9630--9640.

\bibitem{liu2024exploring}
xingbin liu, J.~Zhou, T.~Kong, X.~Lin, and R.~Ji, ``Exploring target representations for masked autoencoders,'' in \emph{Proc. Int. Conf. Learn. Represent. (ICLR)}, May 2024.

\bibitem{bao2021beit}
H.~Bao, L.~Dong, S.~Piao, and F.~Wei, ``{BE}i{T}: {BERT} pre-training of image transformers,'' in \emph{Proc. Int. Conf. Learn. Represent. (ICLR)}, Apr. 2022.

\bibitem{yang2023emoset}
J.~Yang, Q.~Huang, T.~Ding, D.~Lischinski, D.~Cohen-Or, and H.~Huang, ``Emo{S}et: A large-scale visual emotion dataset with rich attributes,'' in \emph{Proc. {IEEE/CVF} Int. Conf. Comput. Vis. (ICCV)}, Oct. 2023, pp. 20\,383--20\,394.

\bibitem{redmon2018yolov3}
J.~Redmon and A.~Farhadi, ``Yolov3: An incremental improvement,'' 2018, \textit{arXiv:1804.02767}.

\bibitem{hinton2015distillingknowledgeneuralnetwork}
G.~Hinton, ``Distilling the knowledge in a neural network,'' 2015, \textit{arXiv:1503.02531}.

\bibitem{mikels2005emotional}
J.~A. Mikels, B.~L. Fredrickson, G.~R. Larkin, C.~M. Lindberg, S.~J. Maglio, and P.~A. Reuter-Lorenz, ``Emotional category data on images from the international affective picture system,'' \emph{Behav. Res. Methods}, vol.~37, no.~4, pp. 626--630, 2005.

\bibitem{machajdik2010affective}
J.~Machajdik and A.~Hanbury, ``Affective image classification using features inspired by psychology and art theory,'' in \emph{Proc. 18th ACM Int. Conf. Multimedia}, Oct. 2010, pp. 83--92.

\bibitem{borth2013large}
D.~Borth, R.~Ji, T.~Chen, T.~Breuel, and S.-F. Chang, ``Large-scale visual sentiment ontology and detectors using adjective noun pairs,'' in \emph{Proc. 21st ACM Int. Conf. Multimedia}, Oct. 2013, pp. 223--232.

\bibitem{peng2015mixed}
K.-C. Peng, T.~Chen, A.~Sadovnik, and A.~C. Gallagher, ``A mixed bag of emotions: Model, predict, and transfer emotion distributions,'' in \emph{Proc. {IEEE} Conf. Comput. Vis. Pattern Recognit. (CVPR)}, Jun. 2015, pp. 860--868.

\bibitem{yang2022emotion}
D.~Yang, S.~Huang, S.~Wang, Y.~Liu, P.~Zhai, L.~Su, M.~Li, and L.~Zhang, ``Emotion recognition for multiple context awareness,'' in \emph{Proc. Eur. Conf. Comput. Vis. (ECCV)}, Oct. 2022, pp. 144--162.

\bibitem{you2016building}
Q.~You, J.~Luo, H.~Jin, and J.~Yang, ``Building a large scale dataset for image emotion recognition: The fine print and the benchmark,'' in \emph{Proc. {AAAI} Conf. Artif. Intell. (AAAI)}, vol.~30, no.~1, Feb. 2016.

\bibitem{lee2019context}
J.~Lee, S.~Kim, S.~Kim, J.~Park, and K.~Sohn, ``Context-aware emotion recognition networks,'' in \emph{Proc. {IEEE/CVF} Int. Conf. Comput. Vis. (ICCV)}, Oct. 2019, pp. 10\,142--10\,151.

\bibitem{ekman1993facial}
P.~Ekman, ``Facial expression and emotion,'' \emph{Am. Psychol.}, vol.~48, no.~4, p. 384, 1993.

\bibitem{dosovitskiy2020image}
A.~Dosovitskiy, L.~Beyer, A.~Kolesnikov, D.~Weissenborn, X.~Zhai, T.~Unterthiner, M.~Dehghani, M.~Minderer, G.~Heigold, S.~Gelly, J.~Uszkoreit, and N.~Houlsby, ``An image is worth 16x16 words: Transformers for image recognition at scale,'' in \emph{Proc. Int. Conf. Learn. Represent. (ICLR)}, May 2021.

\bibitem{simonyan2014very}
K.~Simonyan and A.~Zisserman, ``Very deep convolutional networks for large-scale image recognition,'' 2014, \textit{arXiv:1409.1556}.

\bibitem{he2016deep}
K.~He, X.~Zhang, S.~Ren, and J.~Sun, ``Deep residual learning for image recognition,'' in \emph{Proc. {IEEE} Conf. Comput. Vis. Pattern Recognit. (CVPR)}, Jun. 2016, pp. 770--778.

\bibitem{huang2017densely}
G.~Huang, Z.~Liu, L.~Van Der~Maaten, and K.~Q. Weinberger, ``Densely connected convolutional networks,'' in \emph{Proc. {IEEE} Conf. Comput. Vis. Pattern Recognit. (CVPR)}, Jul. 2017, pp. 2261--2269.

\bibitem{9880205}
Z.~Xie, Z.~Zhang, Y.~Cao, Y.~Lin, J.~Bao, Z.~Yao, Q.~Dai, and H.~Hu, ``Sim{MIM}: a simple framework for masked image modeling,'' in \emph{Proc. {IEEE} Conf. Comput. Vis. Pattern Recognit. (CVPR)}, Jun. 2022, pp. 9643--9653.

\bibitem{10204508}
S.~Ren, F.~Wei, Z.~Zhang, and H.~Hu, ``Tiny{MIM}: An empirical study of distilling {MIM} pre-trained models,'' in \emph{Proc. {IEEE} Conf. Comput. Vis. Pattern Recognit. (CVPR)}, Jun. 2023, pp. 3687--3697.

\bibitem{10205066}
C.~Tao, X.~Zhu, W.~Su, G.~Huang, B.~Li, J.~Zhou, Y.~Qiao, X.~Wang, and J.~Dai, ``Siamese image modeling for self-supervised vision representation learning,'' in \emph{Proc. {IEEE} Conf. Comput. Vis. Pattern Recognit. (CVPR)}, Jun. 2023, pp. 2132--2141.

\bibitem{10204021}
J.~Liu, X.~Huang, J.~Zheng, Y.~Liu, and H.~Li, ``Mix{MAE}: Mixed and masked autoencoder for efficient pretraining of hierarchical vision transformers,'' in \emph{Proc. {IEEE} Conf. Comput. Vis. Pattern Recognit. (CVPR)}, Jun. 2023, pp. 6252--6261.

\bibitem{chen2024context}
X.~Chen, M.~Ding, X.~Wang, Y.~Xin, S.~Mo, Y.~Wang, S.~Han, P.~Luo, G.~Zeng, and J.~Wang, ``Context autoencoder for self-supervised representation learning,'' \emph{Int. J. Comput. Vis.}, vol. 132, no.~1, pp. 208--223, Jan. 2024.

\bibitem{zhang2022visual}
Y.~Zhang, W.~Ding, R.~Xu, X.~Hu, and L.~De~Raedt, ``Visual emotion representation learning via emotion-aware pre-training,'' in \emph{Proc. 31st Int. Joint Conf. Artif. Intell. (IJCAI)}, Jul. 2022, pp. 1679--1685.

\end{thebibliography}

\begin{IEEEbiography}[{\includegraphics[width=1in,height=1.25in,clip,keepaspectratio]{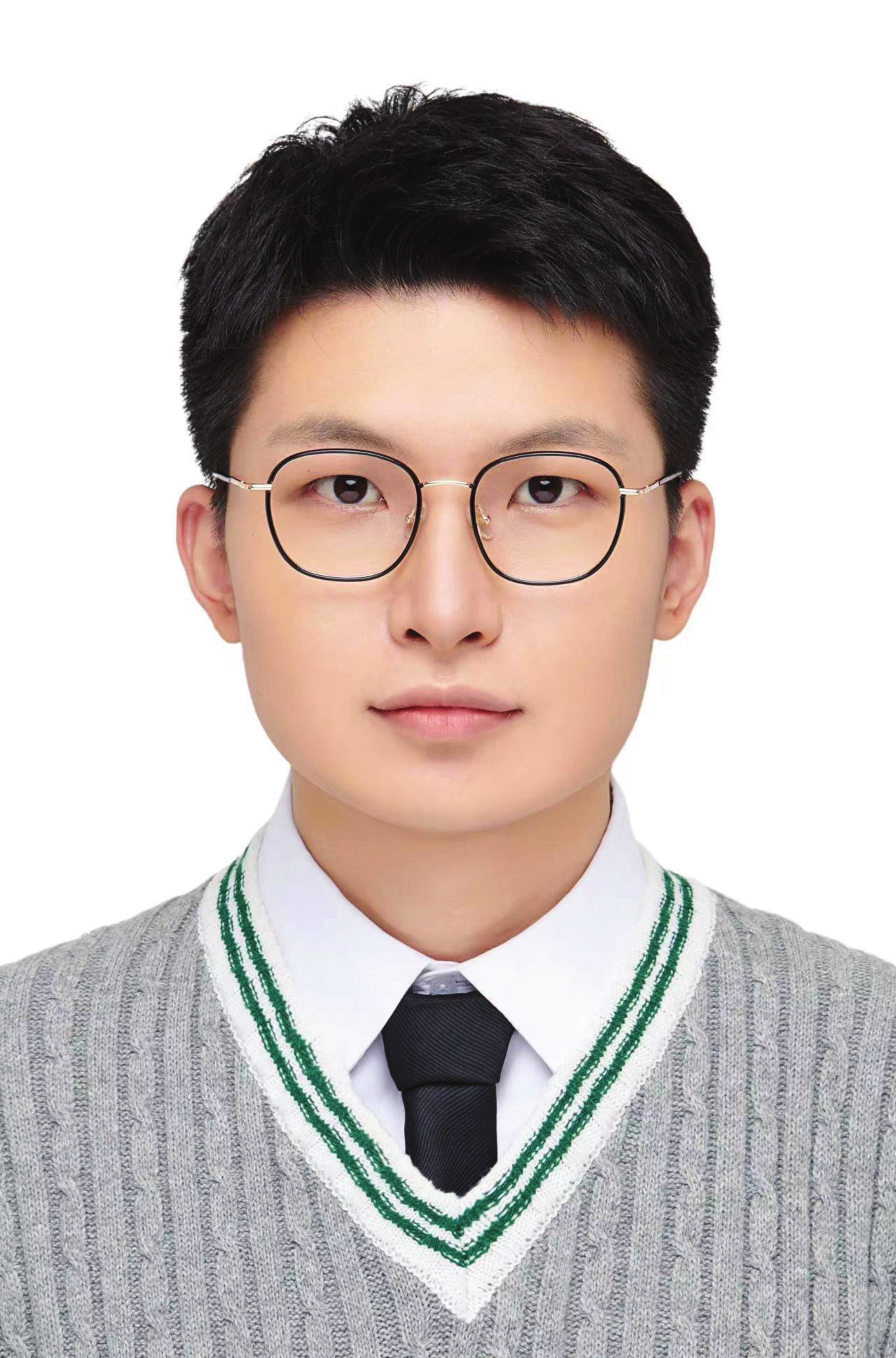}}]{Chuang Chen}
  was born in 1999. He received the M.E. degree from the School of Artificial Intelligence, Anhui University, Hefei, China, in 2024. 
  From 2022 to 2024, he was a Visiting Student with the Institute of Artificial Intelligence, 
  Hefei Comprehensive National Science Center, Hefei, China. 
  His research interests include computer vision and affective computing.
  \end{IEEEbiography}

  \begin{IEEEbiography}[{\includegraphics[width=1in,height=1.25in,clip,keepaspectratio]{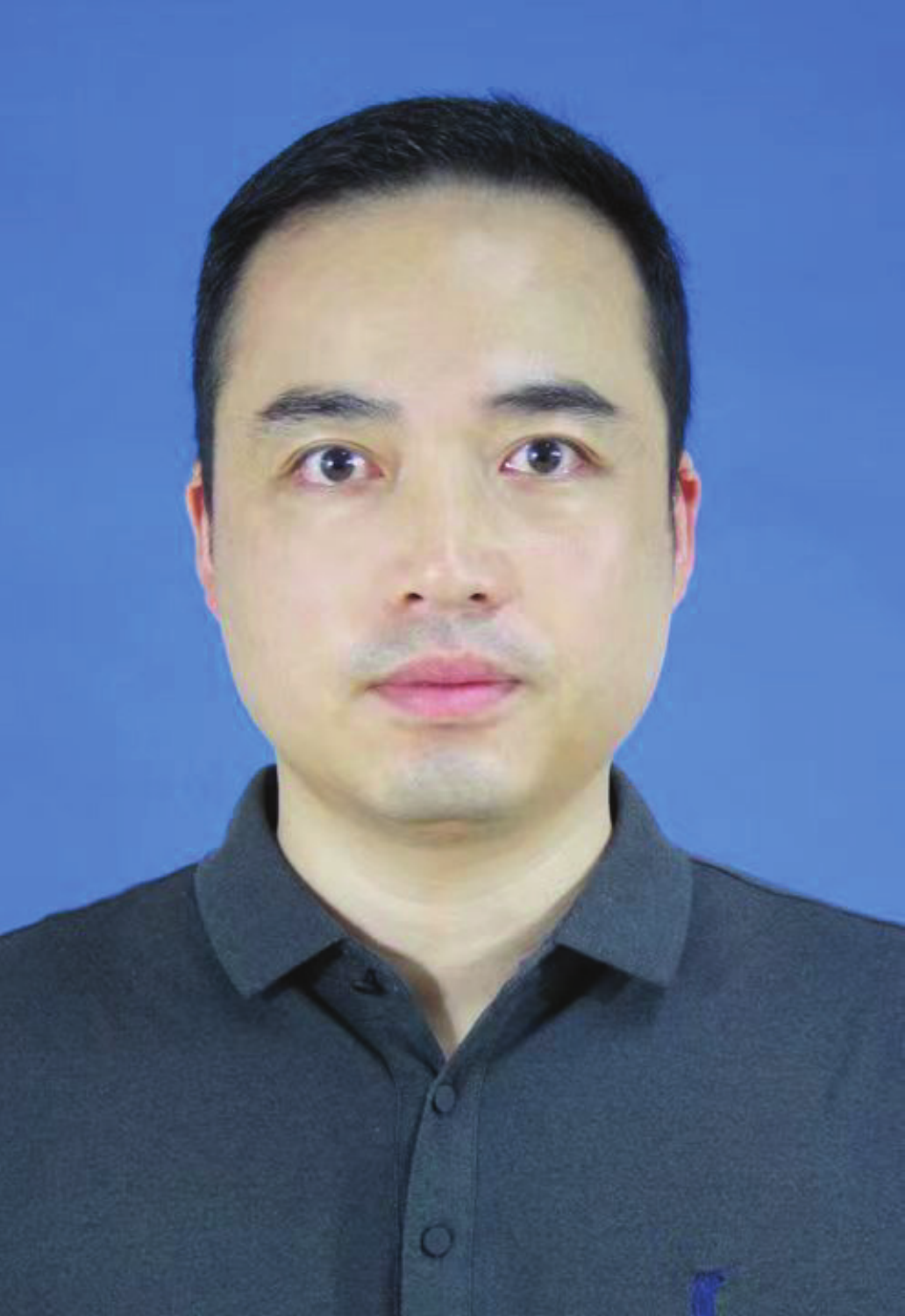}}]{Xiao Sun}
    (Senior Member, IEEE) was born in 1980. He received the M.E. degree from the Department of 
    Computer Sciences and Engineering, Dalian University of Technology, Dalian, China, in 2004, and the dual Ph.D. degree from the University of Tokushima, Tokushima, Japan, in 2009, and the Dalian University of Technology, in 2010. His field of study was natural language processing. 
    He is currently working as a Professor with the Anhui Province Key Laboratory of Affective Computing and Advanced Intelligent Machine, Hefei University of Technology, Hefei, China. 
    His research interests include affective computing, natural language processing, machine learning and human-machine interaction.  
  \end{IEEEbiography}

  \begin{IEEEbiography}[{\includegraphics[width=1in,height=1.25in,clip,keepaspectratio]{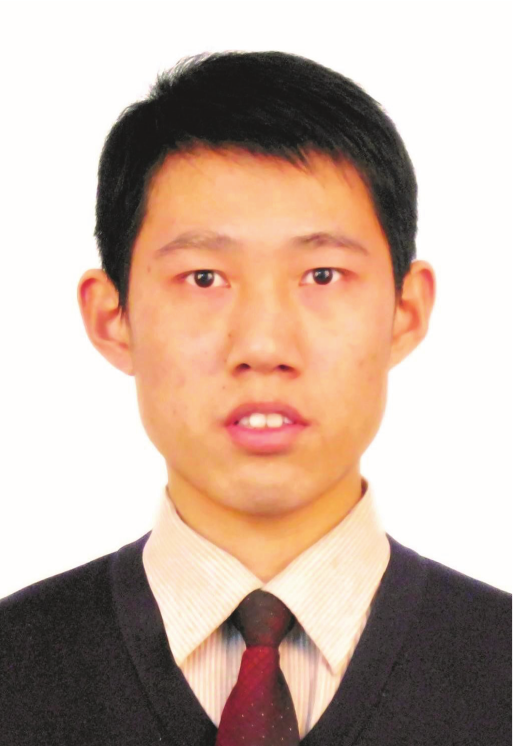}}]{Zhi Liu} 
    (Senior Member, IEEE) received the Ph.D. degree in informatics from the National Institute of
    Informatics, Tokyo, Japan, in 2014. He is currently an Associate Professor with The University of Electro-Communications, Tokyo. 
    His research interests include video network transmission and mobile edge computing.
    He was a recipient of the multiple IEEE best paper awards. He is now an Editorial Board Member of \textit{Wireless Networks} (Springer), 
    \textit{IEEE Transactions on Multimedia} and \textit{IEEE Open Journal of the Communications Society}.
 \end{IEEEbiography}

\vfill

\end{document}